\documentclass{article}
\usepackage{arxiv}

\usepackage[T1]{fontenc}

\usepackage{graphicx}
\usepackage[table]{xcolor}
\usepackage{subcaption}
\usepackage{tikz}
\usetikzlibrary{
  arrows.meta,
  positioning,
  shapes.geometric,
  fit,
  calc,
  backgrounds,
  shadows
}

\usepackage{amsmath}
\usepackage{amsfonts}
\usepackage{amssymb}
\usepackage{amsthm}

\theoremstyle{definition}
\newtheorem{example}{Example}[section]
\usepackage{booktabs}
\usepackage{multirow}
\usepackage{multicol}

\usepackage{algorithm}
\usepackage{algpseudocode}

\usepackage{soul}
\usepackage{todonotes}

\usepackage{hyperref}
\usepackage{cleveref}

\usepackage[acronym]{glossaries}

\newacronym{ml}{ML}{Machine Learning}
\newacronym{mokp}{MOKP}{multiobjective knapsack problem}
\newacronym{mospp}{MOSPP}{multiobjective set packing problem}
\newacronym{motsp}{MOTSP}{multiobjective traveling salesperson problem}

\newcommand{\method}{NOSH} 

\newcommand{\noshRule}{\texttt{NOSH-Rule}}
\newcommand{\noshFE}{\texttt{NOSH-ML-FE}}
\newcommand{\noshEE}{\texttt{NOSH-ML-E2E}}

\tikzset{
  block/.style={
    rectangle, draw, rounded corners,
    text centered, fill=gray!20,
    minimum height=1.2cm
  },
  dashedblock/.style={
    rectangle, dashed, draw, rounded corners,
    text centered,
    minimum height=1.2cm
  },
  line/.style={draw, -{Latex}},
  aggblock/.style={
    rectangle, draw, rounded corners,
    text centered, fill=blue!20,
    minimum height=1cm
  }
}

\title{Heuristic Multiobjective Discrete Optimization using Restricted Decision Diagrams}

\date{} 					

\author{{Rahul Patel$^1$, Elias B. Khalil$^1$, David Bergman$^2$} \\
        $^1$Department of Mechanical and Industrial Engineering,	University of Toronto\\	
        $^2$Department of Operations and Information Management,
	University of Connecticut\\
        \texttt{rm.patel@mail.utoronto.ca, khalil@mie.utoronto.ca} 
}

\renewcommand{\shorttitle}{Heuristic Multiobjective Discrete Optimization using Restricted DDs}

\hypersetup{
pdftitle={Heuristic Multiobjective Discrete Optimization using Restricted Decision Diagrams},
pdfsubject={},
pdfauthor={Rahul~Patel, Elias B.~Khalil, David~Bergman},
pdfkeywords={Multiobjective Optimization, Decision Diagrams, Machine Learning},
}

\begin{document}
\maketitle

\begin{abstract}
Decision diagrams (DDs) have emerged as a state-of-the-art method for exact multiobjective integer linear programming. When the DD is too large to fit into memory or the decision-maker prefers a fast approximation to the Pareto frontier, the complete DD must be restricted to a subset of its states (or nodes). We introduce new node-selection heuristics for constructing restricted DDs that produce a high-quality approximation of the Pareto frontier. Depending on the structure of the problem, our heuristics are based on either simple rules, machine learning with feature engineering, or end-to-end deep learning. Experiments on multiobjective knapsack, set packing, and traveling salesperson problems show that our approach is highly effective, recovering over $85\%$ of the Pareto frontier while achieving $2.5\times$ speedups over exact DD enumeration on average, with very few non-Pareto solutions. The code is available at~\href{https://github.com/rahulptel/HMORDD}{https://github.com/rahulptel/HMORDD}.
\end{abstract}


\section{Introduction}\label{sec:Intro}


Real-world decision-making rarely involves a single criterion. Instead, one must frequently trade off conflicting goals, such as maximizing profit while minimizing risk, or maximizing service quality while minimizing cost. These scenarios are modeled with \textit{multiobjective optimization} (MOO). Unlike single-objective optimization, where the goal is to find a single optimal solution, the goal in MOO is to identify the \textit{Pareto frontier}—the set of feasible solutions for which no objective can be improved without worsening another. 
In this work, we focus on \textit{multiobjective integer linear programming} (MOILP), i.e., MOO problems with integer variables and linear constraints and objectives. Just as the framework of integer linear programming is widely used to model single-objective decision-making tasks, its multiobjective counterpart has found various applications in multicriteria decision-making~\cite{ehrgott2006multicriteria,ehrgott2016exact}. 

In recent years, \textit{decision diagrams} (DDs) have emerged as a powerful alternative to traditional enumerative methods. Initially introduced for representing switching circuits~\cite{lee1959representation} and later popularized for formal verification~\cite{bryant1986graph}, DDs were adapted for discrete optimization problems by compactly encoding their feasible set~\cite{bergman2016decision}.
A DD is a directed acyclic graph where paths from the root to the terminal node correspond to feasible solutions.
Building on this success, recent works have extended DDs to the multiobjective setting \cite{bergman2016multiobjective,bergman2021network} and achieved state-of-the-art results for problems amenable to dynamic programming formulations. By representing the solution space compactly, exact DDs often outperform traditional algorithms for problems with 2--10 objectives and hundreds of variables.

The major bottleneck for exact methods such as DDs is that the size of the Pareto frontier can grow exponentially with instance size, making exact enumeration computationally prohibitive and often overwhelming for the decision-maker. In many practical scenarios, a user requires that a high-quality approximation of the frontier be delivered quickly, rather than a delayed exact set. To address this issue, several heuristics have been proposed in the literature~\cite{deb2002fast,zhang2007moea,tricoire2012multi,pal2019feasibility,pal2019fpbh,an2024matheuristic}. 

Within the context of DDs, one can \textit{restrict} the diagram to obtain an approximate Pareto frontier.  A restricted DD limits the width of the graph (the number of nodes per layer), thereby discarding some feasible solutions to maintain a manageable size. While restricted DDs are widely used to obtain bounds~\cite{bergman2011manipulating,bergman2014optimization,cappart2019improving,nafar2024using} in single-objective problems, their application to multiobjective optimization remains largely unexplored. 
To the best of our knowledge, this work is the first to address the challenge of constructing a restricted DD that quickly recovers a large fraction of the true Pareto frontier, thereby extending the use of DDs to heuristic MOO. 

\paragraph{Illustrative example.} Consider the knapsack problem in \Cref{ex:mokp} and the corresponding exact DD in~\Cref{fig:kp_dds}.
\begin{example}[A bi-objective knapsack problem]
\label{ex:mokp}
\begin{align*}
\min_{x \in \{0, 1\}^3} (x_1 + 10x_2 + 3x_3, ~2x_1 + 3x_2 + x_3)\;\text{such that}\; 3x_1 + x_2 + 2x_3 \leq 5.
\end{align*}
\end{example}
The state in this case, represented by a square node, is equal to the weight of the items selected in the knapsack. We start with an empty knapsack, represented by the root node $\mathbf{r}$ with weight 0. From there on, in each layer we decide whether to select an item or not. For example, selecting (not selecting) item $x_1$ is represented by a solid (dashed) arc from the root node, resulting in a state with a value 3 (or 0). Note that only the red nodes are used by the two Pareto solutions. A restricted DD comprising only of the red nodes preserves the Pareto frontier. Additionally, the red nodes are a small fraction of the total nodes in the DD. Hence, if one can construct a restricted DD with only the red nodes, the Pareto frontier can be recovered quickly.

\begin{figure}[tbp!]
    \centering
    \begin{subfigure}[t]{0.23\textwidth}
        \centering
        \resizebox{\linewidth}{!}{
        \begin{tikzpicture}[level/.style={sibling distance=30mm/#1}, font=\footnotesize]
          \node[rectangle, minimum size=12pt, draw=red, label={$\mathbf r$}, inner sep=0pt] (r) at (0,0) {0};
          \node[rectangle, minimum size=12pt, draw=red, inner sep=0pt] (11) at (-0.5,-1.5) {3};
          \node[rectangle, minimum size=12pt, draw=red, inner sep=0pt] (12) at (0.5,-1.5) {0};
          \node[rectangle, minimum size=12pt, draw=red, inner sep=0pt] (21) at (-1.5,-3) {4};
          \node[rectangle, minimum size=12pt, draw, inner sep=0pt] (22) at (-0.5, -3) {3};
          \node[rectangle, minimum size=12pt, draw=red, inner sep=0pt] (23) at (0.5, -3) {1};
          \node[rectangle, minimum size=12pt, draw, inner sep=0pt] (24) at (1.5,-3) {0};
          \node[rectangle, minimum size=12pt, draw=red, label={below:{$\mathbf t$}}, inner sep=0pt] (t) at (0,-4.5) {/};
          
          \draw[->, red] (r) -- (11);
          \draw[dashed, ->, red] (r) -- (12);  
          \draw[->, red] (11) -- (21);
          \draw[dashed, ->] (11) -- (22);
          \draw[->, red] (12) -- (23);
          \draw[dashed, ->] (12) -- (24);    
          \draw[dashed, ->, red] (21) to[bend right] (t);
          \draw[->] (22) to[bend right] (t);  
          \draw[dashed, ->] (22) -- (t);  
          \draw[->, red] (23) to[bend left] (t);
          \draw[dashed, ->] (23) -- (t);  
          \draw[->] (24) to[bend left] (t);
          \draw[dashed, ->] (24) to[bend left=10] (t);
        \end{tikzpicture}}
        \caption{}
    \end{subfigure}
    \hfill
    \begin{subfigure}[t]{0.10\textwidth}
        \centering
        \resizebox{\linewidth}{!}{
        \begin{tikzpicture}[level/.style={sibling distance=30mm/#1}, font=\footnotesize]
          \node[rectangle, minimum size=12pt, draw=red, label={$\mathbf r$}, inner sep=0pt] (r) at (0,0) {0};
          \node[rectangle, minimum size=12pt, draw=red, inner sep=0pt] (11) at (-0.5,-1.5) {3};
          \node[rectangle, minimum size=12pt, draw=red, inner sep=0pt] (12) at (0.5,-1.5) {0};
          \node[rectangle, minimum size=12pt, draw=red, inner sep=0pt] (21) at (-0.5,-3) {4};
          \node[rectangle, minimum size=12pt, draw=red, inner sep=0pt] (22) at (0.5, -3) {1};
          \node[rectangle, minimum size=12pt, draw=red, label={below:{$\mathbf t$}}, inner sep=0pt] (t) at (0,-4.5) {/};
          
          \draw[->, red] (r) -- (11);
          \draw[dashed, ->, red] (r) -- (12);  
          \draw[->, red] (11) -- (21);
          \draw[->, red] (12) -- (22);
          \draw[dashed, ->, red] (21) to[bend right] (t);
          \draw[->, red] (22) to[bend left] (t);  
          \draw[dashed, ->] (22) -- (t);            
        \end{tikzpicture}}
        \caption{}        
    \end{subfigure}    
    \hfill
    \begin{subfigure}[t]{0.10\textwidth}
        \centering
        \resizebox{\linewidth}{!}{
        \begin{tikzpicture}[level/.style={sibling distance=30mm/#1}, font=\footnotesize]
          \node[rectangle, minimum size=12pt, draw=red, label={$\mathbf r$}, inner sep=0pt] (r) at (0,0) {0};
          \node[rectangle, minimum size=12pt, draw=red, inner sep=0pt] (11) at (-0.5,-1.5) {3};
          \node[rectangle, minimum size=12pt, draw=red, inner sep=0pt] (12) at (0.5,-1.5) {0};
          \node[rectangle, minimum size=12pt, draw=red, inner sep=0pt] (21) at (-0.5,-3) {4};
          \node[rectangle, minimum size=12pt, draw, inner sep=0pt] (22) at (0.5, -3) {3};
          \node[rectangle, minimum size=12pt, draw=red, label={below:{$\mathbf t$}}, inner sep=0pt] (t) at (0,-4.5) {/};
          
          \draw[->, red] (r) -- (11);
          \draw[dashed, ->, red] (r) -- (12);  
          \draw[->, red] (11) -- (21);
          \draw[dashed, ->] (11) -- (22);
          \draw[dashed, ->, red] (21) to[bend right] (t);
          \draw[->] (22) to[bend left] (t);  
          \draw[dashed, ->] (22) -- (t);            
        \end{tikzpicture}}
        \caption{}        
    \end{subfigure}
    \hfill    
    \begin{subfigure}[t]{0.17\textwidth}
        \centering
        \resizebox{\linewidth}{!}{
        \begin{tikzpicture}[level/.style={sibling distance=30mm/#1}, font=\footnotesize]
          \node[rectangle, minimum size=12pt, draw=red, label={$\mathbf r$}, inner sep=0pt] (r) at (0,0) {0};
          \node[rectangle, minimum size=12pt, draw=red, inner sep=0pt] (11) at (-0.5,-1.5) {3};
          \node[rectangle, minimum size=12pt, draw=red, inner sep=0pt] (12) at (0.5,-1.5) {0};
          \node[rectangle, minimum size=12pt, draw, inner sep=0pt] (21) at (-0.5,-3) {3};
          \node[rectangle, minimum size=12pt, draw, inner sep=0pt] (22) at (0.5, -3) {0};
          \node[rectangle, minimum size=12pt, draw=red, label={below:{$\mathbf t$}}, inner sep=0pt] (t) at (0,-4.5) {/};
          \node (13) at (1.5, 0) {$x_1$};
          \node (14) at (1.5, -1.5) {$x_2$};
          \node (15) at (1.5, -3) {$x_3$};
          
          \draw[->, red] (r) -- (11);
          \draw[dashed, ->, red] (r) -- (12);  
          \draw[dashed, ->] (11) -- (21);
          \draw[dashed, ->] (12) -- (22);
          \draw[dashed, ->] (21) to[bend right] (t);
          \draw[->] (21) -- (t);
          \draw[->] (22) to[bend left] (t);  
          \draw[dashed, ->] (22) -- (t);            
        \end{tikzpicture}}
        \caption{}        
    \end{subfigure}
    \caption{
        The exact and restricted DDs for \Cref{ex:mokp}. The paths corresponding to the Pareto-optimal solutions 
        \((1,1,0)\) and \((0,1,1)\), along with their associated Pareto nodes, are highlighted in red. (a) Exact DD. (b) Restricted DD with complete recovery of the Pareto frontier. (c) Restricted DD with partial recovery of the Pareto frontier. (d) Restricted DD with no recovery of the Pareto frontier. 
    }
    \label{fig:kp_dds}
\end{figure}

\begin{table}[b!]
\caption{Average percentage of Pareto nodes (rounded) across problem classes and instance sizes (number of objectives and variables), computed over 10 instances.} 
\label{tab:pareto-state-frac}
\centering
\resizebox{0.75\linewidth}{!}{
\begin{tabular}{c c c c c c}
\toprule
    MOSPP & Pareto Node (\%) & ~~MOKP & Pareto Node (\%) & ~~MOTSP & Pareto Node (\%)\\
\midrule
    (3, 100) & 1 & (3, 80) & 3 & (3, 15) & 2 \\
    (5, 100) & 1 & (5, 40) & 7 & (4, 15) & 6 \\
    (7, 100) & 2 & (7, 40) & 12 & & \\     
\bottomrule
\end{tabular}}
\end{table}

Table \Cref{tab:pareto-state-frac} reports the percentage of Pareto nodes observed across three benchmark problem classes—MOSPP, MOKP, and MOTSP—for varying numbers of objectives and decision variables, averaged over ten instances per configuration. Overall, the results indicate that Pareto nodes constitute only a small fraction of the total nodes in a DD.
\paragraph{Empirical justification.} This phenomenon is not accidental: in random instances of the \gls*{mokp}, \gls*{mospp}, and \gls*{motsp}, only a very small fraction of DD nodes (between 1\% and 12\%) lead to Pareto-optimal solutions, as detailed in \Cref{tab:pareto-state-frac}. We refer to these nodes as~\textit{Pareto nodes}. If one could identify these nodes in advance and restrict the DD to them, the enumeration of the Pareto frontier would be substantially faster, as it dramatically reduces the number of nodes that must be explored.

\paragraph{Contributions.} Building on these conceptual and empirical insights, we tackle the challenge of rapidly generating a high-quality approximation of the Pareto frontier by constructing restricted DDs that retain most Pareto nodes. We develop two families of node-selection heuristics (NOSHs) to guide the restriction process: rule-based and learning-based.~\Cref{fig:nosh_framework} illustrates our methods.

Rule-based heuristics do not require prior knowledge about the distribution of problem instances and operate on each instance independently. For example, in the \gls*{mokp}, the DD state is equal to the knapsack weight. A simple rule is to sort the nodes in each layer by decreasing weight and select only the top-ranked nodes up to the width limit.

In contrast, learning-based heuristics assume access to a sufficiently large collection of training instances, along with their Pareto frontiers, from which a binary classifier can be learned. Nodes that are predicted to be Pareto are then prioritized over the ones that are not. Once trained, the classifier is invoked on each node of the DDs of previously unseen instances that are similar in structure to those seen in training.

We demonstrate that NOSHs effectively identify Pareto nodes, leading to much faster solution enumeration compared to exact DDs and recovering a large fraction of the true Pareto frontier on three representative multiobjective problems: \gls*{mokp}, \gls*{mospp}, and \gls*{motsp}.

\begin{figure}[tbp!]
    \centering
    \resizebox{1.0\linewidth}{!}{
    \begin{tikzpicture}[
      node distance = 1.8cm and 2.8cm,
      box/.style = {draw, rectangle, minimum width=2.8cm,
                    minimum height=0.9cm, align=center},
      >=Stealth
    ]
    
    \node[box] (state) {State: \textcolor{red}{4}};
    \node[box, left=0.5cm of state] (problem) {Problem};
    \node[box, right=0.5cm of state] (layer) {Layer: \textcolor{red}{3}};

    \node[draw, rectangle, inner sep=10pt,
          fit=(problem)(state)(layer)] (common) {};draw=red,

    \node[rectangle, align=center, left=1.8cm of common] (inst) { $\begin{aligned}
    \min_{x \in \{0, 1\}^3} & ~~(x_1 + 10x_2 + 3x_3, ~2x_1 + 3x_2 + x_3) \\
    \text{s.t.}&~~3x_1 + x_2 + 2x_3 \leq 5\\
    \end{aligned}$};
   
    \node[rectangle, 
          align=center, below=0.5cm of inst] (dd) {%
      \begin{tikzpicture}[font=\footnotesize]
        \node[rectangle, minimum size=12pt, draw, label={$\mathbf r$}, inner sep=0pt] (r)  at (0, 0)  {0};
        \node[rectangle, minimum size=12pt, draw, inner sep=0pt]                         (11) at (-0.5,-1.5) {3};
        \node[rectangle, minimum size=12pt, draw, inner sep=0pt]                         (12) at (0.5,-1.5)  {0};
        \node[rectangle, minimum size=12pt, draw, inner sep=0pt, color=red] (21) at (-1.5,-3) {4};
        \node[rectangle, minimum size=12pt, draw, inner sep=0pt] (22) at (-0.5, -3) {3};
        \node[rectangle, minimum size=12pt, draw, inner sep=0pt] (23) at (0.5, -3) {1};
        \node[rectangle, minimum size=12pt, draw, inner sep=0pt] (24) at (1.5,-3) {0};
        \node[] (x1) at (2, 0) {$x_1$};
        \node[] (x2) at (2, -1.5) {$x_2$};
        \draw[->]             (r)  -- (11);
        \draw[dashed, ->]     (r)  -- (12);  
        \draw[->]             (11) -- (21);
        \draw[dashed, ->]          (11) -- (22);
        \draw[->] (12) -- (23);
        \draw[dashed, ->] (12) -- (24);
      \end{tikzpicture}\\[2pt]
      Decision Diagram
    };

    \node[box, below=1.5cm of problem] (rule) {Rule};
    \node[box, below=1.5cm of state] (feat) {Feature\\Engineering};
    \node[draw, rectangle, minimum width=2.8cm, minimum height=1.4cm,
          below=1.5cm of layer, align=center] (nn) {%
      \begin{tikzpicture}[x=1pt,y=1pt]
        \tikzset{neuron/.style={circle, draw, minimum size=4pt, inner sep=0pt}}
    
        \node[neuron] (i1) at (0, 5) {};
        \node[neuron] (i2) at (0,-5) {};
    
        \node[neuron] (h1) at (15,10) {};
        \node[neuron] (h2) at (15, 0) {};
        \node[neuron] (h3) at (15,-10) {};
    
        \node[neuron] (o1) at (30,0) {};
    
        \foreach \i in {i1,i2} {
          \foreach \h in {h1,h2,h3} {
            \draw (\i) -- (\h);
          }
        }
        \foreach \h in {h1,h2,h3} {
          \draw (\h) -- (o1);
        }
      \end{tikzpicture} \\ Neural\\Network
    };
    
    \node[draw, rectangle, minimum width=2.8cm, minimum height=1.4cm,
          below=1cm of feat, align=center] (model) {
      \begin{tikzpicture}[x=1pt,y=1pt]
        \node[circle, draw, minimum size=4pt, inner sep=0pt] (a) at (0,0) {};
        \node[circle, draw, minimum size=4pt, inner sep=0pt] (b) at (-10,-12) {};
        \node[circle, draw, minimum size=4pt, inner sep=0pt] (c) at (10,-12) {};
        \node[circle, draw, minimum size=4pt, inner sep=0pt] (d) at (-14,-24) {};
        \node[circle, draw, minimum size=4pt, inner sep=0pt] (e) at (-6,-24) {};
        \node[circle, draw, minimum size=4pt, inner sep=0pt] (f) at (14,-24) {};
        \node[circle, draw, minimum size=4pt, inner sep=0pt] (g) at (6,-24) {};
    
        \draw (a) -- (b);
        \draw (a) -- (c);
        \draw (b) -- (d);
        \draw (b) -- (e);
        \draw (c) -- (f);
        \draw (c) -- (g);
      \end{tikzpicture} \\ Model
    };

    \node[draw, dashed, rectangle, inner sep=10pt,
          fit=(rule)(feat)(model)(nn)] (common2) {};

    \node[box, below=1cm of model] (score) {Score};

    \draw[->] (inst.east) -- (problem.west);

    \draw[->] (common) -- (rule.north);
    \draw[->] (common) -- (feat);
    \draw[->] (common) -- (nn.north);
    
    \draw[->] (feat) -- (model);
    
    \draw[->] (rule.south) |- (score.west);
    \draw[->] (model) -- (score);
    \draw[->] (nn.south) |- (score.east);


    \coordinate (ddexit) at ($(dd.west)+(-2,0)$);
    \coordinate (midup)  at ($(inst.north west)+(-0.7,0.7)$);
    
    \coordinate (statenorth) at ($(midup -| state.north)$);
    \coordinate (layernorth) at ($(midup -| layer.north)$);
    
    \draw (dd.west) -- (ddexit) -- (midup);
    
    \draw (midup) -- (statenorth);
    \draw (midup) -- (layernorth);
    
    \draw[->] (statenorth) -- (state.north);
    \draw[->] (layernorth) -- (layer.north);
    
    \end{tikzpicture}
    }
    \caption{A schematic illustration of the node-selection pipeline in restricted DD construction with width 2. A problem instance and a DD node---represented by its state and layer---are provided as input to a node-selection heuristic. Heuristics may be rule-based or learning-based; the latter include feature-engineered models and neural networks. The heuristic outputs a score for the node, which guides the construction of the width-restricted DD. The heuristic components depicted in the figure are highlighted with a dashed border.}
    \label{fig:nosh_framework}
\end{figure}

\section{Preliminaries}\label{sec:Prelim}

\subsection{Multiobjective Integer Linear Programming}
We define an MOILP with $K$ objectives as $\mathcal P \equiv \min_x~\{Cx : x \in \mathcal{X} \subseteq \mathbb{Z}^N\}$, where $x$ is the decision vector, $\mathcal{X}$ is the feasible set, $C \in \mathbb{R}^{K \times N}$ is the objective coefficient matrix. Let $\mathcal{Z} = \{Cx : x \in \mathcal{X}\}$ denote the set of feasible objective vectors.
For two vectors $z^1, z^2 \in \mathbb{R}^K$, we say $z^1$ \textit{dominates} $z^2$ (denoted $z^1 \prec z^2$) if $z^1_k \leq z^2_k$ for all $k \in \{1, \cdots, K\}$ and $z^1 \neq z^2$. A vector $z \in \mathcal{Z}$ is \textit{nondominated} if there exists no $z' \in \mathcal{Z}$ such that $z' \prec z$. Solving the MOILP amounts to finding the \textit{Pareto frontier} $\mathcal{Z}^\star = \{ z \in \mathcal{Z} \mid \nexists z' \in \mathcal{Z} ~\text{such that}~ z' \prec z \}$, which is the set of all nondominated vectors.
The set of feasible decision vectors that map to the Pareto frontier is defined as the \textit{efficient set} $\mathcal{X}^\star = \{ x \in \mathcal{X} \mid Cx \in \mathcal{Z}^\star \}$.

\subsection{Decision Diagrams}
A DD for problem $\mathcal{P}$ is a layered directed acyclic graph $\mathcal{B} = (\mathcal{N}, \mathcal{A})$ with node set $\mathcal{N}$ and arc set $\mathcal{A}$. The node set is partitioned into $N+1$ layers, $\mathcal{N} = \mathcal{L}_1 \cup \dots \cup \mathcal{L}_{N+1}$. Layers $\mathcal{L}_1$ and $\mathcal{L}_{N+1}$ contain only the single \emph{root node} $\mathbf{r}$ and \emph{terminal node} $\mathbf{t}$, respectively. The \emph{width} of the DD is the maximum size of any layer, denoted $\omega(\mathcal{B}) = \max_{j} |\mathcal{L}_j|$. For example, the exact DD in \Cref{fig:kp_dds} has 4 layers and a width of 4.

Arcs are directed from layer $j$ to $j+1$ for $j \in \{1, \dots, N\}$. For each arc $a \in \mathcal{A}$ originating from a node in $\mathcal{L}_j$, we define two attributes. An \emph{assignment value} $d(a) \in \mathbb{Z}$, representing the value assigned to decision variable $x_j$. An \emph{objective weight} $v(a) \in \mathbb{R}^K$, representing the contribution of this assignment to the objective function. A path $p = (a_1, \dots, a_N)$ from $\mathbf{r}$ to $\mathbf{t}$ defines a complete solution $x(p) = (d(a_1), \dots, d(a_N)) \in \mathbb{Z}^N$ with an associated objective vector $v(p) = \sum_{i=1}^N v(a_i)$. 
For instance, the path $(\mathbf{r}-3, 3-4, 4-\mathbf{t})$ in the exact DD of \Cref{fig:kp_dds} corresponds to the decision vector $(x_1, x_2, x_3) = (1, 1, 0)$ and achieves an objective of $(11, 5)$.

Let $\mathcal{X}_{\mathcal{B}} = \{ x(p) : p \text{ is an } \mathbf{r}\text{-}\mathbf{t} \text{ path in } \mathcal{B} \}$ be the set of solutions encoded by the DD. The DD is \emph{exact} for problem $\mathcal{P}$ if $\mathcal{X}_{\mathcal{B}} = \mathcal{X}$ and the arc weights satisfy $v(p) = Cx(p)$ for all paths.
Under these conditions, the image of $\mathcal{X}_{\mathcal{B}}$ in objective space is $\mathcal{Z}_{\mathcal{B}} = \{ v(p) : p \text{ is an } \mathbf{r}\text{-}\mathbf{t} \text{ path} \}$. The nondominated vectors in $\mathcal{Z}_{\mathcal{B}}$ are denoted by $\mathcal{Z}^{\star}_{\mathcal{B}}$ and is equal to the Pareto frontier $\mathcal{Z}^\star$ of $\mathcal{P}$.

A decision diagram $\mathcal{B}'$ is said to be \emph{restricted} for problem $\mathcal{P}$ if it encodes a subset of the feasible solutions, i.e., $\mathcal{X}_{\mathcal{B}'} \subseteq \mathcal{X}$. 
In practice, restricted DDs are often maintained by imposing a \emph{maximum width} $W$ on the graph. During construction, if the number of nodes in a layer $\mathcal{L}_j$ exceeds $W$, a heuristic filtering procedure is triggered. This procedure retains a subset of $W$ nodes and removes the rest, ensuring that $\omega(\mathcal{B}') \leq W$.
Consequently, the image of the feasible set $\mathcal{Z}_{\mathcal{B}'}$ is a subset of the objective space $\mathcal{Z}$. The nondominated vectors in $\mathcal{Z}_{\mathcal{B}'}$ are represented by $\mathcal{Z}^{\star}_{\mathcal{B}'}$, which is an approximate Pareto frontier of $\mathcal{P}$.

To construct a DD we leverage the dynamic programming formulation of the problem. As a result, each node $u \in \mathcal N$ is associated with a \emph{state} $s(u)$ from a state space $\mathcal{S}$ given by the formulation. 
The state plays a key role in the design of \method{} for constructing restricted DDs. We refer the reader to~\cite{bergman2016multiobjective,bergman2021network} for additional details on dynamic programming formulation and Pareto frontier enumeration using the multicriteria shortest path algorithm.
The notation used to define MOILP and DDs is summarized in \Cref{sec:notation}.
\section{Methodology}
\label{sec:Method}
The central challenge in solving MOILPs via DDs is the exponential growth of the state space $\mathcal S$ with problem size. To address this, we propose \method{}s, a framework for constructing restricted DDs that prioritize the retention of Pareto nodes, which typically form a small fraction of the total nodes as highlighted in \Cref{tab:pareto-state-frac}. Given a maximum width $W$, our goal is to construct a restricted DD $\mathcal{B}'$ such that $\omega(\mathcal{B}') \leq W$ while maximizing the approximation quality of the Pareto frontier.

We formalize this using a generic node scoring framework described in \Cref{sec:nosh}, and subsequently detail how this framework specializes into rule-based heuristics, classical machine learning with feature engineering, and end-to-end deep learning.

\subsection{Node Selection Heuristics}
\label{sec:nosh}
When the width of a layer $\mathcal{L}_j$ exceeds the maximum width $W$, a filtering procedure must select a size-$W$ subset of nodes to retain. We unify this selection process under a single scoring entity, denoted as the \textit{scorer} $S_{\Theta}$.
The scorer is a function $S_{\Theta}: \mathcal{S} \setminus \{\mathbf{s(r), s(t)}\} \times \{2, \dots, N\} \to [0, 1]$ parameterized by $\Theta$. This function maps the state and layer index of a node $u \in \mathcal{N} \setminus \{\mathbf{r, t}\}$'s , $(s(u), l(u))$ to a score representing the likelihood of $u$ being a Pareto node.

Our objective is to learn (or design) parameters $\Theta$ that maximize the expected recovery of the exact Pareto frontier $\mathcal{Z}^\star$. Let $\mathcal{Z}^\star_{\mathcal{B}'(S_{\Theta})}$ denote the approximate Pareto frontier obtained from a restricted DD constructed using scorer $S_{\Theta}$. The optimization objective is:
\begin{equation}
\label{eq:obj}
\max_{\Theta} \mathbb{E}_{\mathcal{P}} \left[ \frac{|\mathcal{Z}^\star_{\mathcal{B}'(S_{\Theta})} \cap \mathcal{Z}^\star|}{|\mathcal{Z}^\star|} \right],
\end{equation}
where the expectation is taken over a distribution of training instances. An ideal scoring function achieves an objective of 1; scoring functions that eliminate too many Pareto nodes achieve a low score, as many of the returned solutions would be dominated. Based on the structure of $S_{\Theta}$ and the nature of $\Theta$, we categorize \method{}s into three distinct approaches.

\paragraph{\textbf{Rule-based scoring.}}
In this setting, $S_{\Theta}$ is a fixed, deterministic function where $\Theta = \emptyset$ (i.e., there are no learnable parameters). These heuristics rely on domain knowledge and intrinsic properties of the state, as illustrated below for example.
\begin{itemize}
    \item \textit{Scalar States:} If $s(u) \in \mathbb{R}$, natural candidates for the scorer are the state value itself ($S(u) = s(u)$, denoted \texttt{Scal+}) or its negation. For example, the state in~\gls*{mokp} is the total weight of the items that are selected in the state. Since the objectives are in the maximization direction, states that include many items can be thought to be conducive to high-quality solutions, leading to a natural scoring rule.
    \item \textit{Set-based States:} If $s(u)$ is a set, the scorer may utilize the cardinality of the set ($S(u) = |s(u)|$, denoted \texttt{Card+}) or its negation. For example, the state in~\gls*{mospp} is the set of items that can be added to already-selected items without violating any of the packing constraints. The larger this set, the more items one can potentially add down the line, the larger the objective values.
\end{itemize}
While computationally inexpensive, these rules are rigid and their performance sensitive to instance structure.

\paragraph{\textbf{Learning-based scoring with feature engineering.}}
Here, the scorer is a composite function $S_{\Theta}(u) = f_{\Theta}(\psi(s(u), l(u)))$. A fixed extraction function $\psi$ maps the raw state and layer to a hand-crafted feature vector in $\mathbb{R}^d$. These features capture our understanding of the problem. A parameterized binary classification model $f_{\Theta}: \mathbb{R}^d \to [0,1]$ (e.g., Logistic Regressor, Random Forest) maps these features to a probability score. In this context, $\Theta$ represents the parameters of the classifier $f$. This approach adapts to data but is limited by the expressiveness of the feature map $\psi$.

\begin{figure}[t]
    \centering
    \resizebox{0.9\textwidth}{!}{
    \begin{tikzpicture}[
        >={Stealth[length=3mm]},
        font=\sffamily\small,
        data/.style={
            draw=blue!40!black, 
            fill=blue!5, 
            thick, 
            rectangle, 
            rounded corners=3pt, 
            minimum height=1cm, 
            minimum width=2.5cm, 
            align=center,
            drop shadow
        },
        proc/.style={
            draw=orange!60!black, 
            fill=orange!10, 
            thick, 
            rectangle, 
            rounded corners=3pt, 
            minimum height=1cm, 
            minimum width=3cm, 
            align=center,
            drop shadow
        },
        sum/.style={
            circle, 
            draw=black, 
            fill=white, 
            thick, 
            inner sep=0pt, 
            minimum size=6mm
        },
        label/.style={
            font=\bfseries\footnotesize, 
            text=gray!70!black
        }
    ]

    \node[data, text width=2.5cm] (Input) {Problem Instance\\$\mathcal{P}$};

    \node[proc, right=1.5cm of Input] (Aggregator) {Objective Aggregator\\(Deep Sets)};

    \node[proc, below=1cm of Aggregator] (GNN) {Variable Encoder\\(Graph Neural Network)};

    \node[proc, below=2cm of GNN] (StateEnc) {State Encoder};

    \node[data, text width=2.6cm, left=1.56cm of StateEnc] (State) {Node State\\$s(u)$};

    \node[proc, below=1cm of StateEnc] (LayerEnc) {Layer Encoder};

    \node[data, text width=2.6cm, left=1.5cm of LayerEnc] (LayerIdx) {Layer Index\\$l(u)$};

    \node[sum] (Sum) at ($(StateEnc.east)!0.5!(LayerEnc.east) + (1.5,0)$) {\Large$+$};

    \node[proc, right=1cm of Sum] (Head) {Scoring Head\\(MLP)};

    \node[data, right=1cm of Head] (Score) {Node Score\\$S_\Theta(u)$};

    \draw[->, thick] (Input) -- (Aggregator);
    \draw[->, thick] (Aggregator) -- (GNN);
    \draw[->, thick] (Input.south) |- (GNN.west); 
    
    \draw[->, thick] (Input.south) -- (State.north);
    
    \draw[->, thick] (GNN) -- node[right, font=\footnotesize, align=left] {Variable\\Embedding} (StateEnc);

    \draw[->, thick] (State) -- (StateEnc);
    \draw[->, thick] (LayerIdx) -- (LayerEnc);

    \draw[->, thick] (StateEnc) -| (Sum);
    \draw[->, thick] (LayerEnc) -| (Sum);

    \draw[->, thick] (Sum) -- (Head);
    \draw[->, thick] (Head) -- (Score);

    \begin{pgfonlayer}{background}
        \node[fit=(Input)(Aggregator)(GNN), 
              draw=gray!30, dashed, fill=gray!5, rounded corners, inner sep=10pt,
              label={[anchor=north west, inner sep=5pt]north west:\textbf{Phase 1: Instance Embedding}}] {};
              
        \node[fit=(State)(LayerIdx)(Score)(Head)(StateEnc)(LayerEnc)(Sum), 
              draw=gray!30, dashed, fill=gray!5, rounded corners, inner sep=10pt,
              label={[anchor=south west, inner sep=5pt]south west:\textbf{Phase 2: Node Scoring}}] {};
    \end{pgfonlayer}

    \end{tikzpicture}
    }
    \caption{The generic end-to-end node selection heuristic architecture.}
    \label{fig:nosh_arch}
\end{figure}

\paragraph{\textbf{End-to-end Learning-based scoring.}}
To overcome the limitations of manual feature engineering and capture the complex dependencies between objectives, constraints, and decision variables, we propose a generic end-to-end deep learning architecture. The scorer $S_{\Theta}$ maps the raw state and layer definition $(s(u), l(u))$ directly to a score. As illustrated in \Cref{fig:nosh_arch}, this architecture operates in two distinct phases: a computationally intensive \textit{Instance Embedding Phase} performed once per problem instance, and a lightweight \textit{Scoring Phase} performed for each node in the decision diagram.

The instance embedding phase begins with \textit{Objective Aggregator}. Since the order of objectives is arbitrary, it combines information across objectives into a permutation-invariant representation using Deep Sets \cite{zaheer2017deep}. The key result in \cite{zaheer2017deep} establishes that any permutation-invariant function $f$ mapping a set $O$ to a representation can be approximated in the form $f(O) = \rho\left( \sum_{x \in O} \gamma(x) \right)$, where $\gamma$ and $\rho$ are flexible functions (e.g., neural networks).

The \textit{Variable Encoder}, initialized with the aggregated embeddings from the previous step, captures the intricate dependencies among various components of the problem instance. We exploit the fact that the MOILP can be represented as a graph and use a Graph Neural Network (GNN) to process it. This results in \textit{variable embeddings} that encapsulate the instance's structural information. Crucially, the variable embeddings are computed only once per instance, avoiding redundant computations per DD node.

Later, these pre-computed variable embeddings are combined with dynamic state information to obtain a \textit{State Embedding}. Finally, the state embedding is combined with a \textit{Layer Embedding} (derived from the layer index) to obtain a comprehensive \textit{DD Node Embedding}. This embedding is used by the \textit{Scoring Head} to output a scalar score for a DD node.

\subsection{Training Formulation}
For the learning-based approaches, we employ supervised learning. We generate a node-wise training dataset $\mathcal{D}$ derived from $m$ problem instances. For each instance $i$, we use the ground-truth efficient set $\mathcal{X}_i^\star$ to label the nodes in the DD $\mathcal{N}_i$. The dataset is defined as the union of these labeled nodes:
\begin{equation}
    \mathcal{D} = \{ (\mathcal{P}_i, s(u), l(u), y_u) \mid i \in \{1, \dots, m\}, u \in \mathcal{N}_i \},
\end{equation}
where $y_u \in \{0, 1\}$ is the binary label indicating whether $u$ is a Pareto node.

We seek parameters $\Theta^\star$ that minimize the empirical risk over $\mathcal{D}$ using the weighted binary cross-entropy loss:
\begin{equation}
    \Theta^\star \in \arg\min_{\Theta} \frac{1}{|\mathcal{D}|} \sum_{(\cdot, y_u) \in \mathcal{D}} \mathcal{L}_{\text{WBCE}}(S_{\Theta}(s(u), l(u)), y_u),
    \label{eq:erm}
\end{equation}
where $\mathcal{L}_{\text{WBCE}}(p, y) = -\alpha \cdot y \log p - (1-y) \log (1-p)$ and $\alpha \in \mathbb R_+$. By minimizing this loss, $S_{\Theta}$ learns to assign higher probabilities to Pareto nodes.

\subsection{Case Study: Multiobjective Traveling Salesperson Problem}
\label{sec:case_study_motsp}
In the dynamic programming formulation for the \gls*{motsp} \cite{bertsekas2012dynamic}, a node $u$ at layer $j$ represents a partial tour of length $j-1$. The state $s(u)$ is defined as the tuple $s(u) = (V_u, i)$, where $V_u \subseteq \mathcal{V}$ is the set of vertices visited on the path from the root to $u$, and $i \in V_u$ is the index of the last vertex added to the path (i.e., the current city).
At the root node $\mathbf{r}$ (layer 1), the state is $(\{1\}, 1)$, assuming the tour always starts at vertex 1. For a node $u$ with state $(V_u, i)$, a transition to vertex $j$ is valid if $j \notin V_u$. The succeeding state is $(V_u \cup \{j\}, j)$.
The travel costs associated with different edges are represented using a matrix $c$, where $c^k_{ij}$ denotes the cost of edge $(i, j)$ for the $k$-th objective.

\subsubsection{Rule-based Heuristics}
Our rule-based heuristics leverage the edge cost matrices $c^k$ to estimate the quality of a state. Due to the conflicting nature of the $K$ objectives, we first compute the \textit{rank} of each edge for every objective. Let $R \in \mathbb{R}^{N \times N \times K}$ be a tensor where $R_{ijk}$ is the rank of cost $c_{ij}^k$ among all edges for objective $k$.
We derive a scalar score for node $u$ by aggregating these ranks. First, we aggregate across objectives to obtain a single ``cost'' matrix $R^{\text{agg}} \in \mathbb{R}^{N \times N}$. The entry corresponding to edge $(i, j)$ is given by $R^{\text{agg}}_{ij} = \bigoplus_1 \{R_{ijk} : k \in \{1, \dots, K\}\}$, where $\bigoplus_1$ is an aggregation function (e.g., mean, maximum or minimum).

The score for a node $u$ with state $(V_u, i)$ is determined by the potential extensions to the set of unvisited cities $U_u = \mathcal{V} \setminus V_u$. Specifically, the score is computed as $S(u) = \bigoplus\nolimits_2 \{ R^{\text{agg}}_{ij} : j \in U_u \},$ where $\bigoplus_2 \in \{\text{high}, \text{low}\}$ determines whether we prioritize the best or worst immediate extension\footnote{For two ranks $r_1, r_2 \in \mathbb R_+$, rank $r_1$ is higher than $r_2$ if $r_1 < r_2$.}.
We denote these heuristics using the prefix ``\texttt{Ord}'' followed by the aggregation method and the prioritization direction. For instance, \texttt{OrdMeanHigh} aggregates ranks using the mean ($\bigoplus_1$) and scores the node based on the most favorable (higher rank) extension ($\bigoplus_2$). Conversely, a suffix of ``\texttt{Low}'' implies prioritizing edges with lower aggregated ranks (worst extensions).

\subsubsection{Learning-based Heuristic}

For the learning-based approach, we instantiate the end-to-end architecture proposed in \Cref{fig:nosh_arch}. 
In the instance embedding phase, the input to the objective aggregator are the node features $h_i^{v, 0} \in \mathbb{R}^{N\times d_{node}}$ edge features $h_{ij}^{e, 0} \in \mathbb{R}^{N\times N \times K}$.
The node features are initialized with the 2D coordinates of the cities and may optionally include hand-crafted features. The edge features are initialized using the cost matrices across the $K$ objectives, where $h_{ijk}^{e, 0} = c_{ij}^k$. The objective aggregator transforms these features $h^{v, 0}$ and $h^{e, 0}$ into objective-invariant embeddings $h^{v, 1} \in \mathbb{R}^{N \times d_{\text{emb}}}$ and $h^{e, 1} \in \mathbb{R}^{N \times N \times d_{\text{emb}}}$
 
To capture the complex structural dependencies of the TSP, we utilize an Edge-augmented Graph Transformer (EGT) \cite{hussain2022global}. The EGT extends the standard Transformer by introducing \emph{edge channels} that evolve structurally aware edge embeddings alongside node embeddings.
Let $h^{v, \ell-1} \in \mathbb{R}^{N \times d_{\text{emb}}}$ and $h^{e, \ell-1} \in \mathbb{R}^{N \times N \times d_{\text{emb}}}$ denote the node and edge embeddings input to layer $\ell$. We first linearly transform these embeddings using learnable weight matrices. For a single attention head with dimension $d_k$, we compute the query $Q^{\ell}= h^{v, \ell-1} W_Q^{\ell}$, key $K^{\ell} = h^{v, \ell-1} W_K^{\ell}$, value $V^{\ell} = h^{v, \ell-1} W_V^{\ell}$, structural bias $E^{\ell}=h^{e, \ell-1} W_E^{\ell}$, and gate $G^{\ell} = h^{e, \ell-1} W_G^{\ell}$ matrices, where $W_Q^{\ell}, W_K^{\ell}, W_V^{\ell} \in \mathbb{R}^{d_{\text{emb}} \times d_k}$ are the node projection matrices and $W_E^{\ell}, W_G^{\ell} \in \mathbb{R}^{d_{\text{emb}} \times 1}$ are the edge projection matrices. Consequently, the resulting matrices have dimensions $Q^{\ell}, K^{\ell}, V^{\ell} \in \mathbb{R}^{N \times d_k}$ and $E^{\ell}, G^{\ell} \in \mathbb{R}^{N \times N}$.
The edge-augmented attention $A^{\ell} \in \mathbb{R}^{N \times N}$ is computed by modulating the standard dot-product similarity with the edge-derived bias and gating terms as follows:
\begin{align}
    A^{\ell} &= \mathrm{softmax} \left(\hat{H}^{\ell}\right) \odot \sigma(G^{\ell}),\\
    \hat{H}^{\ell} &= \mathrm{clip}\left(\frac{Q^{\ell} (K^{\ell})^T}{\sqrt{d_k}}\right) + E^{\ell},
\end{align}
where $\sigma(\cdot)$ is the sigmoid function and $\odot$ denotes element-wise multiplication. Here, $E^{\ell}$ acts as a structural bias added to the node affinities, while $G^{\ell}$ gates the attention values, controlling the information flow based on edge attributes. 

The node embeddings are updated by aggregating $V^{\ell}$ weighted by $A^{\ell}$. In practice, this operation is performed by multiple heads in parallel, and their outputs are combined and projected to form the input for the next layer. We stack $L$ such layers to obtain the final instance-specific embeddings $h^{v, L}$ which acts as the variable embedding $h^v_{emb} \in \mathbb R^{N\times d_{emb}}$.

The primary challenge in the node scoring phase is to efficiently map the dynamic DD state $s(u)=(V_u, i)$ to a vector representation using the pre-computed embeddings $h_{\text{emb}}^v$. We represent the set of visited nodes $V_u$ by modifying the variable embeddings. We define a state mask $M \in \mathbb{R}^{N \times d_{\text{emb}}}$ with learnable parameters $\theta_{\text{vis}}$ and $\theta_{\text{unvis}}$. The mask for vertex $j$ is set to $\theta_{\text{vis}}$ if $j \in V_u$ and $\theta_{\text{unvis}}$ otherwise.
This mask is added element-wise to the variable embeddings: $h_{\text{mask}}^v = h_{\text{emb}}^v + M$. The resulting set of embeddings is then aggregated into a single vector $h_{\text{vis}} \in \mathbb{R}^{d_{\text{emb}}}$ using a Deep Sets-based network~\cite{zaheer2017deep}, capturing the composition of the partial tour regardless of visit order. To explicitly capture the current position in the graph, we extract the embedding corresponding to the last visited vertex $i$: $h_{\text{last}} = h_{\text{emb}}^v[i]$. 

Finally, the current layer index $l(u)$ is projected to an embedding $h_{\text{layer}} \in \mathbb{R}^{d_{\text{emb}}}$ via a \textit{Layer Encoder}, a multi-layered perceptron (MLP).
The DD node representation is the sum of these embeddings $h_{\text{node}} = h_{\text{vis}} + h_{\text{last}} + h_{\text{layer}}$.
This vector is passed to the scoring head (an MLP) to predict the likelihood of node $u$ leading to a Pareto-optimal solution.

\section{Computational Setup}
\label{sec:setup}

We test the proposed approach on \gls*{mospp}, \gls*{mokp} and \gls*{motsp}.
We use a computing cluster with Intel Xeon CPU E5-2683 CPUs, a memory limit of 16GB, and a time limit of 1800 seconds. 
The DD manipulation code responsible for generating exact, restricted, or reduced DDs is based on \cite{bergman2021network}.
We create a Python binding for this C++ code using pybind11.
The problem definition and instance-generation scheme are detailed in \Cref{app:prob_def}. 

For the learning-based \method{}, each problem class and instance size has 1000 training, 100 validation, and 100 testing instances. We compute the Pareto frontier using the exact DD and extract all Pareto nodes. As shown in \Cref{tab:pareto-state-frac}, Pareto nodes constitute only a small fraction of total nodes, leading to a significant class imbalance. To address this during dataset construction, we apply undersampling to the negative class: for each Pareto (positive) node, we randomly sample one non-Pareto (negative) node, resulting in a 1:1 ratio between positive and negative samples. This not only creates a balanced dataset but only also limits its size, making the training tractable. 

\textbf{Node Selection Heuristics and Baselines}
As highlighted in \Cref{sec:Method}, NOSHs can be categorized into three distinct types. We refer to the rule-based heuristic as \noshRule{}, the learning-based approach that relies on feature engineering as \noshFE{} and the approach that learns the scoring function end-to-end using deep learning as \noshEE{}.
The \noshEE{} is the most extensible method to other problem classes among the proposed NOSHs. By modeling the MOILP as a graph, modern deep learning architectures can be leveraged to learn rich representations for scoring DD nodes. However, this expressivity incurs the computational cost of training deep learning models with multiple hyperparameters. Consequently, we adopt a complexity-aware strategy: we prioritize the lightweight \noshRule{}, employing learning-based variants only when the problem complexity necessitates it. The state definition along with the implementation details of various NOSHs is provided in \Cref{app:nosh}.
The approach for selecting the width for restricted DDs is given in \Cref{sec:width_selection}.


The learning-based NOSH approach for the \gls*{mokp} leverages XGBoost 2.0.1 \cite{xgboost}, a highly efficient implementation of gradient-boosted decision trees. The primary motivation for choosing XGBoost lies in its ability to deliver strong performance with minimal tuning when meaningful, hand-crafted node features are available. Compared to neural network-based methods, this allows for more interpretable and computationally efficient models. We perform a grid search to select the best-performing model based on validation accuracy, which is then used for evaluation on the test set. The selected models achieve classification accuracies between 85\% and 88\% using a threshold of $\tau = 0.5$ on the predicted scores. The corresponding mean absolute errors fall within the range of 0.18 to 0.23. In contrast, designing hand-crafted features for the \gls*{motsp} is considerably more challenging. Therefore, to develop an end-to-end node scoring model for this setting, we employ a graph transformer implemented using PyTorch \cite{paszke2019pytorch}. Details of the hyperparameters used for the node scorers are provided in \Cref{sec:hyperparams}.

Our primary baseline is the state-of-the-art exact DD method (referenced as Exact) based on coupled enumeration \cite{bergman2021network}. We also benchmarked against NSGA-II. Despite its widespread use, NSGA-II failed to produce high-quality approximations of the Pareto frontier, even with increased runtime. When the population size was scaled to match that of the Exact method, it struggled to find feasible solutions. With smaller, more typical population sizes, the resulting frontier was consistently of lower quality than those obtained by \method{}-based approaches. Consequently, we focus our analysis on the DD-based comparisons. Detailed experimental settings and the supplementary NSGA-II analysis are available in \Cref{app:add_results}.

\textbf{Metrics} We report the following metrics to evaluate and compare the performance of different methods.
\begin{enumerate}
    \item \textbf{Width:} The width of the constructed DDs. Ideally, smaller restricted DDs are preferred, provided they still capture most of the Pareto frontier.
    \item \textbf{Time:} The total time (in seconds) required to construct the DD and enumerate the Pareto frontier. Faster methods are desirable, especially when approximating the frontier for large or complex instances. Note that the time required to compile the DD is a small fraction of the total time, even for the learning-based \method{}. Therefore, we report only the total time for constructing the DD and enumerating the Pareto frontier.
    \item \textbf{Cardinality:} The fraction of true Pareto-optimal solutions captured by a method, relative to the total number of Pareto-optimal solutions. Let $\hat{\mathcal{Z}}^\star$ denote the set of solutions obtained by a given method. Then, cardinality is computed as $(|\hat{\mathcal{Z}}^\star \cap \mathcal{Z}^\star| / |\mathcal{Z}^\star|) \times 100$.
    A higher cardinality indicates that a larger portion of the true Pareto frontier is recovered.
    \item \textbf{Precision:} The fraction of solutions identified by the method that are truly Pareto-optimal. This is calculated as $(|\hat{\mathcal{Z}}^\star \cap \mathcal{Z}^\star|/|\hat{\mathcal{Z}}^\star|)\times 100$. A higher precision suggests that the approximate Pareto frontier consists predominantly of true Pareto-optimal solutions.
    \item \textbf{Inverted Generational Distance (IGD) \cite{coello2004study}:} The average distance from each solution in $\mathcal{Z}^\star$ to its closest solution in $\hat{\mathcal{Z}}^\star$. Lower IGD values indicate a closer and more comprehensive approximation of the true Pareto frontier. To ensure scale invariance across objectives, we normalize each objective using the minimum and maximum values observed in $\mathcal{Z}^\star$ before computing IGD.
\end{enumerate}


\begin{table}[tb!]
    \caption{\gls*{mospp} results averaged over ``Inst.'' test instances. 
    Each column corresponds to a specific instance size $(N,K)$, with rows giving 
    the metrics for the Exact and \noshRule{} methods. Refer to \Cref{sec:setup} 
    for column description.}
    \centering
    \footnotesize
    \resizebox{\linewidth}{!}{
    \begin{tabular}{llrrrrrrrrrr}
\toprule
& & \multicolumn{5}{c}{$N = 100$} & \multicolumn{5}{c}{$N = 150$}\\
\cmidrule(lr){3-7}\cmidrule(lr){8-12}
Metric & Method & $K=3$ & $K=4$ & $K=5$ & $K=6$ & $K=7$ & $K=3$ & $K=4$ & $K=5$ & $K=6$ & $K=7$ \\
\midrule
\multirow{2}{*}{Width} 
  & \textcolor{gray}{Exact} 
  & \textcolor{gray}{5,766} & \textcolor{gray}{6,034} & \textcolor{gray}{5,936} & \textcolor{gray}{5,976} & \textcolor{gray}{5,707} 
  & \textcolor{gray}{471,602} & \textcolor{gray}{518,556} & \textcolor{gray}{464,330} & \textcolor{gray}{468,787} & \textcolor{gray}{590,908} \\
  & \noshRule{} & 50 & 50 & 50 & 50 & 50 & 5,000 & 5,000 & 5,000 & 5,000 & 5,000 \\
\midrule
\multirow{2}{*}{Time $\downarrow$} 
  & \textcolor{gray}{Exact} 
  & \textcolor{gray}{1} & \textcolor{gray}{1} & \textcolor{gray}{2} & \textcolor{gray}{5} & \textcolor{gray}{31} 
  & \textcolor{gray}{11} & \textcolor{gray}{51} & \textcolor{gray}{261} & \textcolor{gray}{567} & \textcolor{gray}{783} \\
  & \noshRule{} & 1 & 1 & 1 & 2 & 13 & 1 & 7 & 77 & 183 & 314 \\
\midrule
\multirow{2}{*}{Cardinality $\uparrow$} 
  & \textcolor{gray}{Exact} 
  & \textcolor{gray}{100} & \textcolor{gray}{100} & \textcolor{gray}{100} & \textcolor{gray}{100} & \textcolor{gray}{100} 
  & \textcolor{gray}{100} & \textcolor{gray}{100} & \textcolor{gray}{100} & \textcolor{gray}{100} & \textcolor{gray}{100} \\
  & \noshRule{} & 85 & 84 & 89 & 87 & 87 & 99 & 99 & 99 & 99 & 99 \\
\midrule
\multirow{2}{*}{Precision $\uparrow$} 
  & \textcolor{gray}{Exact} 
  & \textcolor{gray}{100} & \textcolor{gray}{100} & \textcolor{gray}{100} & \textcolor{gray}{100} & \textcolor{gray}{100} 
  & \textcolor{gray}{100} & \textcolor{gray}{100} & \textcolor{gray}{100} & \textcolor{gray}{100} & \textcolor{gray}{100} \\
  & \noshRule{} & 89 & 90 & 94 & 94 & 93 & 100 & 100 & 99 & 100 & 100 \\
\midrule
\multirow{2}{*}{IGD $\downarrow$} 
  & \textcolor{gray}{Exact} 
  & \textcolor{gray}{0.000} & \textcolor{gray}{0.000} & \textcolor{gray}{0.000} & \textcolor{gray}{0.000} & \textcolor{gray}{0.000} 
  & \textcolor{gray}{0.000} & \textcolor{gray}{0.000} & \textcolor{gray}{0.000} & \textcolor{gray}{0.000} & \textcolor{gray}{0.000} \\
  & \noshRule{} 
    & 0.012 & 0.016 & 0.013 & 0.017 & 0.019 
    & 0.000 & 0.000 & 0.001 & 0.001 & 0.001 \\
\midrule
\multirow{2}{*}{$|\hat{\mathcal{Z}}^\star|$} 
  & \textcolor{gray}{Exact} 
  & \textcolor{gray}{238} & \textcolor{gray}{1,117} & \textcolor{gray}{4,765} & \textcolor{gray}{9,117} & \textcolor{gray}{25,457} 
  & \textcolor{gray}{787} & \textcolor{gray}{6,099} & \textcolor{gray}{29,061} & \textcolor{gray}{59,951} & \textcolor{gray}{103,489} \\
  & \noshRule{} & 226 & 1,051 & 4,591 & 8,550 & 24,534 & 786 & 6,089 & 28,953 & 59,724 & 103,275 \\
\midrule
Inst. &  & 100 & 100 & 100 & 100 & 100 & 100 & 100 & 92 & 54 & 27 \\  
\bottomrule
\end{tabular}}
    \label{tab:mis_result}
\end{table}

\begin{table}[tbp!]
    \caption{MOKP results averaged across 100 test instances. Refer to \Cref{sec:setup} for column description.}
    \centering
    \footnotesize
    \resizebox{0.75\linewidth}{!}{
    \begin{tabular}{rrlrrrrrr}
    \toprule
    $N$ & $K$ & ~~Method & ~~Width & ~~Time $\downarrow$ & ~~Cardinality $\uparrow$ & ~~Precision $\uparrow$ & ~~IGD $\downarrow$ & ~~$|\hat{\mathcal{Z}}^\star|$ \\
    \midrule

    \multirow{6}{*}{40} & \multirow{6}{*}{7} 
        & \textcolor{gray}{Exact} 
        & \textcolor{gray}{9,709} 
        & \textcolor{gray}{84} 
        & \textcolor{gray}{100} 
        & \textcolor{gray}{100} 
        & \textcolor{gray}{0.000} 
        & \textcolor{gray}{25,098} \\
    \cmidrule{3-9}
        & & \multirow{2}{*}{\noshRule{}} 
            & 2,000 & \textbf{2} & 19 & 64 & 0.128 & 4,131 \\
        & &  
            & 3,000 & 19 & 61 & 89 & 0.047 & 12,972 \\
    \cmidrule{3-9}
        & & \multirow{2}{*}{\noshFE{}} 
            & 2,000 & 16 & 60 & 74 & 0.042 & 20,482 \\
        & &  
            & 3,000 & 36 & \textbf{88} & \textbf{96} & \textbf{0.012} & 22,267 \\

    \midrule

    \multirow{6}{*}{50} & \multirow{6}{*}{4} 
        & \textcolor{gray}{Exact} 
        & \textcolor{gray}{12,359} 
        & \textcolor{gray}{7} 
        & \textcolor{gray}{100} 
        & \textcolor{gray}{100} 
        & \textcolor{gray}{0.000} 
        & \textcolor{gray}{3,564} \\
    \cmidrule{3-9}
        & & \multirow{2}{*}{\noshRule{}} 
            & 2,500 & \textbf{1} & 17 & 36 & 0.085 & 1,132 \\
        & &  
            & 3,500 & 2 & 52 & 70 & 0.032 & 2,256 \\
    \cmidrule{3-9}
        & & \multirow{2}{*}{\noshFE{}} 
            & 2,500 & 3 & 61 & 70 & 0.020 & 3,062 \\
        & &  
            & 3,500 & 4 & \textbf{88} & \textbf{92} & \textbf{0.006} & 3,367 \\
    
    \midrule
    
    \multirow{6}{*}{80} & \multirow{6}{*}{3} 
        & \textcolor{gray}{Exact} 
        & \textcolor{gray}{20,097} 
        & \textcolor{gray}{27} 
        & \textcolor{gray}{100} 
        & \textcolor{gray}{100} 
        & \textcolor{gray}{0.000} 
        & \textcolor{gray}{2,442} \\
    \cmidrule{3-9}
        & & \multirow{2}{*}{\noshRule{}} 
            & 4,000 & \textbf{3} & 10 & 19 & 0.064 & 1,015 \\
        & &  
            & 6,000 & 7 & 62 & 71 & 0.013 & 1,954 \\
    \cmidrule{3-9}
        & & \multirow{2}{*}{\noshFE{}} 
            & 4,000 & 6 & 46 & 53 & 0.012 & 2,039 \\
        & &  
            & 6,000 & 12 & \textbf{93} & \textbf{95} & \textbf{0.002} & 2,363 \\

    \bottomrule
    \end{tabular}}
    \label{tab:kp_result}
\end{table}

\section{Results}
\label{sec:Results}
Evidently, some restricted DDs are better than others in that they approximate the true Pareto frontier more completely. We will show how NOSHs finds such ``accurate'' restricted DDs for \gls*{mospp}, \gls*{mokp} and \gls*{motsp} in \Cref{sec:ResultsMOSPP}, \Cref{sec:ResultsMOKP} and \Cref{sec:ResultsMOTSP}, respectively. 
The metrics Width, Cardinality, Precision and $|\hat{\mathcal{Z}}^\star|$ are rounded to the nearest integer, whereas Time is rounded up. In all tables, the Exact method is shown in gray as a reference baseline and is not included in the comparative evaluation, as it represents the ground-truth Pareto frontier.

\subsection{Multiobjective Set Packing Problem}
\label{sec:ResultsMOSPP}

Table~\ref{tab:mis_result} reports the performance of Exact and \noshRule{} for the \gls*{mospp}.
The method \noshRule{} is based on the \texttt{Card+} heuristic detailed in \Cref{sec:nosh}.
The \noshRule{} has Cardinality and Precision in the range of $\sim\!85\%$ to $99\%$ and $\sim\!90\%$ to $99\%$, respectively.
\noshRule{} achieves up to $11\times$ speedup over Exact, 
with most instances exhibiting $2\times$--$7\times$ improvements.

For larger instances (\(N = 150\) and \(K \geq 5\)), the reported metrics are averaged over fewer than 100 test instances. This is because the Exact method often exceeds the memory limit as instance size increases, making it unable to compute the Pareto frontier for all cases. Consequently, we apply \noshRule{} only to those instances where Exact completes within the time and memory constraints, which explains the value of ``Inst.'' being less than 100.
Note that applying learning-based \method{} would be a challenge in this setting as it would depend on labeled training data, which limits its applicability to larger instances as we do not have access to the exact Pareto frontier.
In summary, \noshRule{} achieved a good trade-off between solution quality and enumeration time, without the need for expensive data labeling.

\subsection{Multiobjective Knapsack Problem}
\label{sec:ResultsMOKP}

\begin{table}[tbp!]
    \caption{\gls*{motsp} results averaged across 100 test instances. Refer to \Cref{sec:setup} for column description.}
    \centering
    \footnotesize
    \resizebox{0.75\linewidth}{!}{
    \begin{tabular}{rrlrrrrrr}
    \toprule
    $N$ & $K$ & ~~Method & ~~Width & ~~Time $\downarrow$ & ~~Cardinality $\uparrow$& ~~Precision $\uparrow$& ~~IGD $\downarrow$& ~~$|\hat{\mathcal{Z}}^\star|$ \\
    \midrule
    \multirow{6}{*}{15} 
      & \multirow{3}{*}{3} & \textcolor{gray}{Exact} 
    & \textcolor{gray}{24,024} 
    & \textcolor{gray}{3} 
    & \textcolor{gray}{100} 
    & \textcolor{gray}{100} 
    & \textcolor{gray}{0.000} 
    & \textcolor{gray}{868} \\
      &                    & \noshRule{} & 4,804  & \textbf{2}  & 1 & 2 & 0.085 &   439 \\
      &                    & \noshEE{}   & 4,804  & \textbf{2}  & \textbf{91} & \textbf{95} & \textbf{0.003} &   832 \\
    \cmidrule{2-9}
      & \multirow{3}{*}{4} 
        & \textcolor{gray}{Exact} 
    & \textcolor{gray}{24,024} 
    & \textcolor{gray}{28} 
    & \textcolor{gray}{100} 
    & \textcolor{gray}{100} 
    & \textcolor{gray}{0.000} 
    & \textcolor{gray}{9,210} \\
      &                    
        & \noshRule{} & 4,804  & \textbf{3}  & 1 & 4 & 0.082 &  3,254 \\
      &                    
        & \noshEE{}   & 4,804  & 13 & \textbf{89} & \textbf{95} & \textbf{0.004} &  8,546 \\
    \bottomrule
    \end{tabular}}  
    \label{tab:tsp_result}
\end{table}

The performance of different approaches for the \gls*{mokp} is presented in \Cref{tab:kp_result}. The \noshRule{} method, derived from the \texttt{Scal+} heuristic (\Cref{sec:nosh}), is sensitive to the width of the restricted DD, with Cardinality and Precision ranging from 52\%–62\% and 69\%–89\%, respectively. 
In contrast, \noshFE{} consistently outperforms \noshRule{}, achieving Cardinality between 87\%–92\% and Precision between 92\%–95\%, along with lower IGD values. 
Moreover, \noshFE{} attains speedups of $1.75\times$–$2.33\times$ over Exact in most settings. Overall, these results highlight the advantage of \noshFE{} in achieving superior trade-offs between solution quality and runtime. Additionally, the use of XGBoost with handcrafted features enhances interpretability by enabling identification of the most influential features. 

\subsection{Multiobjective Traveling Salesperson Problem}
\label{sec:ResultsMOTSP}

The results for \gls*{motsp} are presented in Table~\ref{tab:tsp_result}. 
The \noshRule{} method, based on the \texttt{OrdMeanLow} heuristic, achieves significant speedups but performs poorly in terms of solution quality, with Cardinality and Precision close to zero and relatively high IGD values, indicating a weak approximation of the Pareto frontier. 
In contrast, \noshEE{} substantially improves solution quality, achieving Cardinality between 89\%–91\% and Precision around 95\%, along with very low IGD values. 
It also attains speedups of $1.5\times$–$2.15\times$ compared to the Exact method. 
These results demonstrate that \noshEE{} is effective in learning useful node representations from raw data and accurately identifying Pareto-optimal nodes.

\section{Related Work}\label{sec:Related}

As surveyed in \cite{ehrgott2016exact,ehrgott2006multicriteria}, the literature on exact approaches to MOILPs is vast and generally partitioned into
decision space methods~\cite{adelgren2022branch} and objective space methods~\cite{boland2014triangle,boland2015criterion}. The former searches in the space of feasible solutions whereas the latter search in the space of objective vectors. Many of these approaches are confined to biobjective and triobjective problems. Notable exceptions include the KSA algorithm~\cite{kirlik2014new} and the DD approach~\cite{bergman2016multiobjective}. The latter has been shown to outperform KSA by a substantial margin on combinatorial problems that are amenable to a dynamic programming formulation, which is why we focus on this promising algorithmic paradigm in this work. Specifically, our approach allows efficient state-space exploration, similar to \cite{kuroiwa2023large}, by eliminating states less likely to contribute to a Pareto optimal solution. 
Note that the authors of~\cite{bergman2016multiobjective} enhance a basic decision diagram approach in~\cite{bergman2021network} by introducing a series of Pareto frontier preserving operations.  Those would directly apply here, but we utilize the basic decision diagram approach for transparency and ease of implementation.

Evolutionary or genetic algorithms (GAs) have long been used for multiobjective optimization. The Pymoo paper and software package~\cite{blank2020pymoo} summarize and implement state-of-the-art GAs such as NSGA-II. Note that the NSGA-II ~\cite{deb2002fast} is widely used (45,000+ citations) so outperforming it is a good sanity check of the promise of any new method. However, it is known that GAs typically struggle with integer variables and hard constraints, a limitation that is not exhibited by the DD approach. Another class of heuristics based on integer and linear programming appear in~\cite{pal2019feasibility,pal2019fpbh,an2024matheuristic}. They extended the single-objective ``Feasibility Pump''~\cite{fischetti2005feasibility} heuristic to the multiobjective case. The method in~\cite{an2024matheuristic} is evaluated only on triobjective problems with binary variables whereas~\method{} will be applied to problems with more objective and discrete variables, a substantial generalization.  With the expansive literature on heuristic approaches to MOILP, we opted to compare only with NSGA-II as it is the best known and has been a strong competitor and benchmark comparison algorithm for over two decades.  

Relative to single-objective optimization, there has been much less work on ML for multiobjective discrete optimization. ML-based methods have been proposed for unconstrained continuous multiobjective problems that arise in deep learning applications such as multi-task learning~\cite{navon2020learning} or molecule generation~\cite{jain2023multi,lin2022pareto}. Because they are not equipped to deal with hard constraints, these methods do not apply to combinatorial optimization. 
More relevant to this paper is the work of~\cite{NEURIPS2022_710aae91} who train a \textit{graph neural network} (GNN) to guide the exact algorithm of~\cite{tamby2021enumeration}. This GNN-based method is evaluated on knapsack problems with 3-5 objectives only and requires a much larger amount of time than~\method{}. For example, on instances with 4 objectives and 50 variables,~\method{} runs in about 11 seconds on average (\Cref{tab:kp_result}) whereas the method in~\cite{NEURIPS2022_710aae91} (Table 1) runs for 1,000 seconds on a faster CPU than ours. Additionally, no publicly available code was provided for this rather sophisticated GNN architecture, making a direct comparison challenging.
\section{Conclusion and Future Work}
\label{sec:Conclusion}
We demonstrated the use of restricted DDs as a heuristic to solve multiobjective integer linear programming problems. 
In fact, to the best of our knowledge, we are the first to invoke restricted DDs in the context of multiobjective optimization as they have only appeared in the single objective case.
We presented two types of node selection heuristics, rule-based and learning-based, to construct the restricted DDs for the \gls*{mokp}, \gls*{mospp} and \gls*{motsp}. 
The results demonstrate that node selection heuristics provide a high-quality approximation of the true Pareto frontier and are significantly faster than the exact DDs.
Specifically, \noshRule{}, \noshFE{} and \noshEE{} performed exceedingly well for the \gls*{mospp}, \gls*{mokp} and \gls*{motsp}, respectively.

Furthermore, instead of classifying a node in isolation, one can formulate the problem of constructing the restricted DDs as a structured output prediction task. Specifically, given an exact DD the goal is to predict a subgraph consisting only of nodes and arcs used by the Pareto optimal solutions. Predicting a subgraph is a combinatorial task which has received significant attention from the machine learning community~\cite{joachims2009predicting} but has not been applied to optimization applications such as ours here. 
One limitation of our approach is its reliance on the complete Pareto frontier to generate the training dataset for the learning-based node selection heuristics. In the future, we aim to relax this requirement by utilizing partial Pareto frontiers for training data generation. This relaxation may necessitate an increase in the number of training instances to maintain performance.

\begin{appendix}




\section{Mathematical Notation}
\label{sec:notation}
The notation used to describe MOILP and DDs is presented in \Cref{tab:notation_moilp} and \Cref{tab:notation_dd}, respectively.

\begin{table}[htbp!]
\centering
\caption{Mathematical notation for MOILP}
\small
\setlength{\tabcolsep}{6pt}
\resizebox{0.7\textwidth}{!}{
\begin{tabular}{ll}
\toprule
\textbf{Symbol} & \textbf{Description} \\
\midrule
$\mathcal{P}$ & A multiobjective integer linear programming problem \\
$N$ & Number of decision variables \\
$K$ & Number of objectives \\
$x \in \mathbb{Z}^N$ & Integer-valued decision vector \\
$\mathcal{X} \subseteq \mathbb{Z}^N$ & Feasible solution set \\
$C \in \mathbb{R}^{K \times N}$ & Objective function coefficient matrix \\
$\mathcal{Z}$ & Set of feasible objective vectors (Image of $(\mathcal{X})$ in the objective space) \\
$\prec$ & Dominance relation: $z^1 \prec z^2$ implies $z^1$ dominates $z^2$ \\
$\mathcal{Z}^\star \subseteq \mathcal{Z}$ & Pareto frontier (set of nondominated objective vectors) \\
\bottomrule
\end{tabular}}
\label{tab:notation_moilp}
\end{table}

\begin{table}[htbp!]
\centering
\caption{Mathematical notation for DDs}
\small
\setlength{\tabcolsep}{6pt}
\resizebox{0.6\textwidth}{!}{
\begin{tabular}{ll}
\toprule
\textbf{Symbol} & \textbf{Description} \\
\midrule
$\mathcal{B} = (\mathcal{N}, \mathcal{A})$ & A decision diagram for problem $\mathcal{P}$ \\
$\mathcal{N}$ & Set of nodes in a DD \\
$\mathcal{A}$ & Set of arcs (directed edges) in a DD \\
$\mathcal{L}_j$ & Set of nodes in layer $j$, for $j \in \{1, \dots, N+1\}$ \\
$\mathbf{r}, \mathbf{t}$ & Root node (layer 1) and terminal node (layer $N+1$) \\
$\omega(\mathcal{B})$ & Width of the DD: $\max_{j} |\mathcal{L}_j|$ \\
$a \in \mathcal{A}$ & An arc in the DD \\
$d(a)$ & Assignment value of arc $a$ (value for variable $x_j$) \\
$v(a)$ & Objective weight of arc $a$ (vector in $\mathbb{R}^K$) \\
$p$ & A path from $\mathbf{r}$ to $\mathbf{t}$ \\
$x(p)$ & Solution vector encoded by path $p$ \\
$v(p)$ & Objective vector accumulated along path $p$ \\
$\mathcal{X}_\mathcal{B}$ & Set of feasible integer solutions encoded by $\mathcal{B}$ \\
$\mathcal{Z}_\mathcal{B}$ & Set of objective vectors encoded by $\mathcal{B}$ \\
$\mathcal{Z}^{\star}_{\mathcal{B}}$ & Set of nondominated vectors in $\mathcal{Z}_\mathcal{B}$ (Pareto frontier) \\
\midrule
\multicolumn{2}{l}{\textit{Restricted DDs and Construction}} \\
\midrule
$\mathcal{B}'$ & Restricted DD where $\mathcal{X}_{\mathcal{B}'} \subseteq \mathcal{X}$ \\
$W$ & Maximum width parameter for restricted DDs \\
$\mathcal{Z}^{\star}_{\mathcal{B}'}$ & Approximate Pareto frontier derived from $\mathcal{B}'$ \\
$\mathcal{S}$ & State space defined by the DP formulation of $\mathcal P$ \\
$s(u) \in \mathcal S$ & State associated with node $u$ \\
\bottomrule
\end{tabular}}
\label{tab:notation_dd}
\end{table}

\section{Problem Definition and Instance Generation}
\label{app:prob_def}

In this section, we provide the formulation of the MOILP problems considered in this work and corresponding instance generation scheme.

\subsection{Multiobjective Travelling Salesperson Problem}
\subsubsection{Problem Definition} The \gls*{motsp} extends the classical traveling salesperson problem by incorporating multiple, potentially conflicting cost metrics associated with traversing arcs. We consider a complete directed graph $\mathcal{G} = (\mathcal{V}, \mathcal{E})$ where $\mathcal{V} = \{1, \ldots, N\}$ is the set of vertices (cities) and $\mathcal{E} = \{(i, j) : i, j \in \mathcal{V}, i \neq j\}$ is the set of edges. 
There are $K$ cost matrices, where $c^k_{ij}$ denotes the cost of edge $(i, j)$ for the $k$-th objective. The problem consists of finding a Hamiltonian cycle (a permutation of the vertices) that simultaneously minimizes the objective vectors. In the context of the MOILP formulation, the decision variables $x = (x_1, \dots, x_N) \in \mathcal{V}^N$ represent the permutation of vertices visited, such that $x_j$ is the index of the vertex visited at step $j$.

\subsubsection{Instance Generation}
The instances are generated as in \cite{ozpeynirci2010exact}. For each $K$, we generated integer coordinates for $N$ cities on a $1000 \times 1000$ square (uniformly at random) and used Euclidean distances to create the distance matrix.

\subsection{Multiobjective Knapsack Problem}
\subsubsection{Problem Definition} The \gls*{mokp} involves selecting a subset of items to maximize multiple profit objectives subject to a single weight capacity constraint. 
Given $N$ items, let $w \in \mathbb{Z}^N_+$ be the vector of item weights and $B \in \mathbb{Z}_+$ be the knapsack capacity. The profit of item $j$ for the $k$-th objective is given by the entry $C_{kj}$.
The problem is formulated as:
\[
\max_{x \in \{0, 1\}^N} \left\{ Cx \mid \sum_{j=1}^N w_j x_j \leq B \right\}.
\]
Here, the decision variable $x_j=1$ implies item $j$ is selected, and $x_j=0$ otherwise.

\subsubsection{Instance Generation}
We follow the instance generation scheme used in \cite{kirlik2014new}, where each profit $C_{kj}$ and weight $w_j$ were drawn uniformly at random from the integer interval $[1, \dots, 1000]$. The capacity of the knapsack was set to $B := \lceil 0.5 \sum_{j=1}^N w_j \rceil$.

\subsection{Multiobjective Set Packing Problem}
\subsubsection{Problem Definition} The \gls*{mospp} seeks to select a subset of variables to maximize objectives subject to pairwise conflict constraints (or packing constraints). Let $A \in \{0, 1\}^{M \times N}$ be a binary constraint matrix with $M$ constraints, and $\mathbf{1}$ be a vector of ones of size $M$. The problem is defined as:
\[
\max_{x \in \{0, 1\}^N} \left\{Cx : Ax \leq \mathbf{1} \right\}.
\]
The constraint $Ax \leq \mathbf{1}$ implies that for any row $m$ of $A$, at most one variable $j$ with $A_{mj}=1$ can be set to 1. This is equivalent to finding a maximum weight independent set on the intersection graph defined by $A$.

\subsubsection{Instance Generation}
The instance generation for the \gls*{mospp} is based on the previous work by \cite{stidsen2014branch}. Specifically, we fix the number of constraints $M = N/5$. Then for each constraint, we select the number of variables that participate in it from a random uniform distribution over [2, 20], resulting in an average of 10 variables per constraint.

We identified a minor issue in this instance generation process. Specifically, certain variables were not involved in any constraints. To address this, we randomly assign such variables to any one of the existing constraints, ensuring all variables participate meaningfully in the problem formulation. 
\section{Implementation Details of Node Selection Heuristics}
\label{app:nosh}

In this section, we describe the implementation of the node selection heuristics for the \gls*{mokp} and \gls*{mospp}. The corresponding details for the \gls*{motsp} are presented in \Cref{sec:case_study_motsp}.

\subsection{Multiobjective knapsack problem}
\label{sec:nosh_mokp}

\subsubsection{State Definition}
The state $s(u)$ at layer $j$ (where the decision for item $j-1$ has just been made) tracks the resource consumption. It is defined as a scalar:
\begin{equation}
    s(u) = q,
\end{equation}
where $q \in \mathbb{Z}_{\ge 0}$ represents the accumulated weight of items selected in the path from the root to node $u$.
Formally, if $u$ is reached by a path $p$ with assignments $x_1, \dots, x_{j-1}$, then $s(u) = \sum_{t=1}^{j-1} w_t x_t$.
A transition $x_j=1$ is feasible only if $s(u) + w_j \leq B$. If $x_j=1$, the next state is $s(u) + w_j$; if $x_j=0$, the next state is $s(u)$.

\subsubsection{Rule-based Heuristic}
As the state in \gls*{mokp} is scalar-valued, the \method{}-Rule method uses Scal+ to select the nodes along with minWeight variable ordering \cite{patel2024leo}. In minWeight variable ordering, we sort the items based on the ascending order of their weight before constructing the DD. 

\subsubsection{Learning-based Heuristic} The features used by the \method{}-ML method are presented in \Cref{tab:kp_features}. Note that these features rely on the fact that the problem has only one constraint. Extending this approach to problems with multiple constraints can be non-trivial. 

\begin{table*}[htbp]
\caption{Features associated with a DD node for the \gls*{mokp}.}
\label{tab:kp_features}
\centering
\resizebox{\textwidth}{!}{
\begin{tabular}{p{4cm} p{12cm} p{1cm}}
    \toprule
    \textbf{Feature scope} & \textbf{Feature} & \textbf{Count} \\
    \midrule
    \multirow{7}{*}{Instance features} & The number of objectives & 1\\
    \cmidrule{2-3}
    & The number of items (or variables) & 1\\
    \cmidrule{2-3}
    & The capacity of the knapsack & 1\\
    \cmidrule{2-3}
    & The mean, min., max., and std. of the weights & 4\\
    \cmidrule{2-3}
    & The mean, min., max., and std. of the values & 12 \\
    \midrule
    \multirow{7}{*}{Layer-variable features} & The weight associated with variable & 1\\
    \cmidrule{2-3}
    & Average value across objectives  & 1\\ 
    \cmidrule{2-3}
    & Maximum value across objectives & 1\\ 
    \cmidrule{2-3}        
    & Minimum value across objectives & 1\\
    \cmidrule{2-3}
    & Standard deviation across objectives & 1\\
    \cmidrule{2-3}
    & Ratio of average value across objectives to weight & 1\\ 
    \cmidrule{2-3}
    & Ratio of maximum value across objectives to weight & 1\\ 
    \cmidrule{2-3}
    & Ratio of minimum value across objectives to weight & 1\\ 
    \midrule
    Layer-index features & Normalized layer index & 1 \\
    \midrule
    State features & Normalized state (weight of the knapsack at the current node) & 1 
    \\
    \cmidrule{2-3}
    & Ratio of state by capacity & 1 \\
    \midrule        
    \textbf{Total} & & 44 \\  
    \bottomrule
\end{tabular}
}    
\end{table*}

\subsection{Multiobjective set packing problem}
\label{sec:nosh_mospp}
\subsubsection{State Definition}
The state $s(u)$ at layer $j$ must capture the availability of the remaining variables $x_j, \dots, x_N$ given the decisions made so far. The state is defined as the set of \emph{eligible} variables:
\begin{equation}
    s(u) = \mathcal{V}_{eligible} \subseteq \{j, \dots, N\},
\end{equation}
where $k \in \mathcal{V}_{eligible}$ implies that selecting variable $k$ (setting $x_k=1$) does not violate any constraints given the variables already selected in the path to $u$.
At the root node, $s(\mathbf{r}) = \{1, \dots, N\}$. 
Given a node $u$ at layer $j$ with state $s(u)$:
\begin{itemize}
    \item If $x_j = 0$, the constraint set remains unchanged for the remaining variables. The next state is $s(u) \setminus \{j\}$.
    \item If $x_j = 1$ (valid only if $j \in s(u)$), we must remove $j$ and all future variables $k > j$ that conflict with $j$ (i.e., variables $k$ such that $\exists m, A_{mj}=1 \land A_{mk}=1$). The next state is $\{ k \in s(u) \setminus \{j\} \mid \text{variable } k \text{ does not conflict with } j \}$.
\end{itemize}

\subsubsection{Rule-based Heuristic}
The \method{}-Rule method uses the Card+ heuristic to select states as they are represented as a set. Ties among nodes with the same cardinality are broken by randomly selecting a subset of them to fit the width limit. To build the DD, we use the minState \cite{bergman2016decision} variable ordering, where we select the variable appearing in the minimum number of states in the current layer to construct the next layer.

\section{Additional Computational Details}
\label{app:setup}

\subsection{Width Selection for Restricted DDs}
\label{sec:width_selection}
One of the key considerations in constructing restricted DDs is determining its width. If the width is too small, the DD may be overly restrictive, making it difficult to recover the true Pareto frontier. Conversely, if the width is too large, the performance of the restricted DD may closely resemble that of the exact DD, offering little computational advantage. 

In the DD literature, the width is typically chosen based on empirical validation, tuned to the downstream task performance using the restricted DD.
In this work, we begin by setting the initial width to approximately 20\% of the average width of the exact DDs for a given problem class and instance size. If the method achieves both cardinality and precision above 80\%, we retain the current width. Otherwise, we incrementally increase the width budget by 10\%. Conversely, if the method performs exceptionally well with the initial width, we systematically reduce it to identify a minimal effective width.

As demonstrated in \Cref{sec:Results}, effective width budgets for the \gls*{mospp}, \gls*{mokp}, and \gls*{motsp} are found to be 1\%, 30\%, and 20\% of the average exact DD width, respectively.

\subsection{Hyperparameter Configuration}
\label{sec:hyperparams}
\Cref{sec:hyperparams_gtf} provides the architectural and training details for the learning-based heuristic for the \gls*{motsp}. \Cref{sec:hyperparams_xgb} outlines the configuration of the learning-based heuristic for the \gls*{mokp}, which is based on XGBoost. Finally, \Cref{sec:hyperparams_nsga} describes the parameters of the NSGA-II-based evolutionary baseline.

\subsubsection{MOTSP scorer configuration}
\label{sec:hyperparams_gtf}
We enrich the raw vertex features of the \gls*{motsp} by computing additional features based on the distances per objective. Specifically, for each vertex, we compute the mean, minimum, maximum, standard deviation, and interquartile range (75th percentile minus 25th percentile) of the distances to other nodes for each objective. These five statistical features are concatenated with the original 2D coordinates of the vertex, resulting in a 7-dimensional feature vector for each vertex. This enhanced representation provides the model with richer contextual information about each node's relative position and importance in the instance.

The Objective Aggregator processes this input using a Deep Sets architecture, which is permutation invariant over objectives. It transforms the enriched node features \( h_0^v \in \mathbb{R}^{N \times K \times 7} \) and edge features \( h_0^e \in \mathbb{R}^{N \times N \times K} \) into fixed-size embeddings \( h_{\text{agg}}^v \in \mathbb{R}^{N \times 32} \) and \( h_{\text{agg}}^e \in \mathbb{R}^{N \times N \times 32} \), respectively, with an embedding dimension of 32. These embeddings are then passed into the downstream Graph Transformer network.

Concretely, we define feature encoders and aggregators using functions:
\[
\gamma_v: \mathbb{R}^{7} \rightarrow \mathbb{R}^{32}, \quad \rho_v: \mathbb{R}^{64} \rightarrow \mathbb{R}^{32}, \quad 
\gamma_e: \mathbb{R} \rightarrow \mathbb{R}^{32}, \quad \rho_e: \mathbb{R}^{32} \rightarrow \mathbb{R}^{32}
\]
Here, \( \gamma_v \) and \( \gamma_e \) are per-objective encoders applied independently to each node and edge for a given objective, while \( \rho_v \) and \( \rho_e \) are permutation-invariant aggregators that combine the representations across objectives. The output of this aggregation serves as a unified embedding that captures information across all objectives. These encoders and aggregators are implemented as a single linear layer with ReLU activation in the output. 

The EGT is configured with 4 layers and 8 heads per layer and no dropout. The architecture uses ReLU as activation and omits biases in both multi-head attention and linear layers. A hidden-to-input dimension ratio of 2 is used in the MLP blocks. The model is trained for 100 epochs using the Adam optimizer. The learning rate is linearly warmed up from \(5 \times 10^{-5}\) to \(5 \times 10^{-4}\), after which it follows a cosine decay schedule down to a minimum of \(5 \times 10^{-5}\).

\subsubsection{MOKP scorer configuration}
\label{sec:hyperparams_xgb}
\begin{table}[tb!]
\centering
\small 
\setlength{\tabcolsep}{6pt} 
\renewcommand{\arraystretch}{1.1} 
\caption{Hyperparameter configuration for the XGBoost-based scorers used in \gls*{mokp}}
\begin{tabular}{lccc}
\toprule
 & \multicolumn{3}{c}{\textbf{Size}} \\
 \cmidrule{2-4}
\textbf{Parameter} & (7, 40) & (4, 50) & (3, 80) \\
\midrule
{max\_depth} & 9 & 7 & 5 \\
{min\_child\_weight} & 10000 & 10000 & 1000 \\
{eta} & \multicolumn{3}{c}{0.3} \\
{objective} & \multicolumn{3}{c}{binary:logistic} \\
{num\_round} & \multicolumn{3}{c}{250} \\
{early\_stopping\_rounds} & \multicolumn{3}{c}{20} \\
{eval\_metric} & \multicolumn{3}{c}{logloss} \\
\bottomrule
\end{tabular}
\label{tab:hyperparams_xgb}
\end{table}

\Cref{tab:hyperparams_xgb} describes the configuration details for the XGBoost-based node scorer for \gls*{mokp}, including the number of trees, learning rate, and maximum depth.

\begin{table}[t!]
    \centering
    \footnotesize
    \caption{NSGA-II Configuration}
    \begin{tabular}{rrr}
        \toprule
        Config & MOKP \& MOSPP & MOTSP \\
        \midrule
        Crossover & TwoPointCrossover & OrderCrossover\\
        Mutation & BitflipMutation & InversionMutation\\
        Sampling & ~~~BinaryRandomSampling & ~~~PermutationRandomSampling\\
        \bottomrule
    \end{tabular}
    \label{tab:nsga_config}
\end{table}

\subsubsection{NSGA II Configuration}
\label{sec:hyperparams_nsga}
The crossover, mutation, and sampling strategies for the NSGA-II algorithm applied to different problems is provided in \Cref{tab:nsga_config}. For each problem type and size, we set the time budget for \gls*{mospp}, \gls*{mokp}, and \gls*{motsp} to match the average runtime of \noshRule{}, \noshFE{}, and \noshEE{}, respectively. Initially, we set the population size to the average number of nondominated solutions observed for each problem size. However, this led to out-of-memory errors for some sizes, prompting us to reduce the population size accordingly.

\section{Additional results}
\label{app:add_results}

\Cref{tab:mis_result_complete}, \Cref{tab:kp_result_complete}, and \Cref{tab:tsp_result_complete} present the complete NSGA-II results for \gls*{mospp}, \gls*{mokp}, and \gls*{motsp}, respectively. In addition, \Cref{tab:tsp_result_complete} includes the performance of various rule-based \method{} variants explored in our experiments.

\begin{table}[htbp!]
    \caption{\gls*{motsp} results averaged across test instances.
    Methods prefixed with \texttt{Ord} correspond to different rule-based NOSHs.
    NSGA-II-$p$ denotes NSGA-II with population size $p$.
    Refer to \Cref{sec:setup} for column description.}
    \centering
    \footnotesize
    \resizebox{0.7\linewidth}{!}{
    \begin{tabular}{rrlrrrrrr}
    \toprule
    $N$ & $K$ & ~~Method & ~~Width & ~~Time $\downarrow$ & ~~Cardinality $\uparrow$& ~~Precision $\uparrow$& ~~IGD $\downarrow$& ~~$|\hat{\mathcal{Z}}^\star|$ \\
    \midrule
    \multirow{20}{*}{15} 
      & \multirow{10}{*}{3} 
        & \textcolor{gray}{Exact} 
    & \textcolor{gray}{24,024} 
    & \textcolor{gray}{3} 
    & \textcolor{gray}{100} 
    & \textcolor{gray}{100} 
    & \textcolor{gray}{0.000} 
    & \textcolor{gray}{868} \\
      &  & NSGA-II-100 & -- & 3 & 4 & 31 & 3.558 & 100 \\
      &  & NSGA-II-500 & -- & 3 & 7 & 20 & 3.531 & 278 \\
      &  & \texttt{OrdMeanHigh} & 4,804 & 2 & 0 & 1 & 0.116 & 376 \\
      &  & \texttt{OrdMeanLow}  & 4,804 & 2 & 1 & 2 & 0.085 & 439 \\
      &  & \texttt{OrdMaxHigh}  & 4,804 & 2 & 0 & 1 & 0.116 & 366 \\
      &  & \texttt{OrdMaxLow}   & 4,804 & 2 & 1 & 3 & 0.090 & 425 \\
      &  & \texttt{OrdMinHigh}  & 4,804 & 2 & 0 & 1 & 0.105 & 369 \\
      &  & \texttt{OrdMinLow}   & 4,804 & \textbf{1} & 1 & 2 & 0.087 & 440 \\
      &  & \noshEE{}            & 4,804 & 2 & \textbf{91} & \textbf{95} & \textbf{0.003} & 832 \\
    \cmidrule{2-9}
      & \multirow{10}{*}{4} 
        & \textcolor{gray}{Exact} 
    & \textcolor{gray}{24,024} 
    & \textcolor{gray}{28} 
    & \textcolor{gray}{100} 
    & \textcolor{gray}{100} 
    & \textcolor{gray}{0.000} 
    & \textcolor{gray}{9,210} \\
      &  & NSGA-II-100 & -- & 25 & 0 & 11 & 4.056 & 100 \\ 
      &  & NSGA-II-500 & -- & 25 & 2 & 28 & 4.008 & 500 \\ 
      &  & \texttt{OrdMeanHigh} & 4,804 & \textbf{2} & 0 & 2 & 0.096 & 2,737 \\
      &  & \texttt{OrdMeanLow}  & 4,804 & \textbf{2} & 1 & 4 & 0.082 & 3,254 \\
      &  & \texttt{OrdMaxHigh}  & 4,804 & \textbf{2} & 1 & 2 & 0.094 & 2,721 \\
      &  & \texttt{OrdMaxLow}   & 4,804 & \textbf{2} & 1 & 3 & 0.081 & 3,281 \\
      &  & \texttt{OrdMinHigh}  & 4,804 & \textbf{2} & 0 & 1 & 0.095 & 2,790 \\
      &  & \texttt{OrdMinLow}   & 4,804 & \textbf{2} & 1 & 2 & 0.084 & 3,229 \\
      &  & \noshEE{}            & 4,804 & 7 & \textbf{89} & \textbf{95} & \textbf{0.004} & 8,546 \\
    \bottomrule
    \end{tabular}}  
    \label{tab:tsp_result_complete}
\end{table}

\begin{table}[htbp!]
    \caption{\gls*{mospp} results averaged over test instances.
    Each column corresponds to a specific instance size $(N,K)$.
    NSGA-II-$p$ denotes NSGA-II with population size $p$.
    Refer to \Cref{sec:setup} for column description.}
    \centering
    \footnotesize
    \resizebox{\linewidth}{!}{
    \begin{tabular}{llrrrrrrrrrr}
\toprule
& & \multicolumn{5}{c}{$N = 100$} & \multicolumn{5}{c}{$N = 150$}\\
\cmidrule(lr){3-7}\cmidrule(lr){8-12}
Metric & Method & $K=3$ & $K=4$ & $K=5$ & $K=6$ & $K=7$ & $K=3$ & $K=4$ & $K=5$ & $K=6$ & $K=7$ \\
\midrule

\multirow{4}{*}{Width} 
  & \textcolor{gray}{Exact} & \textcolor{gray}{5,766} & \textcolor{gray}{6,034} & \textcolor{gray}{5,936} & \textcolor{gray}{5,976} & \textcolor{gray}{5,707} & \textcolor{gray}{471,602} & \textcolor{gray}{518,556} & \textcolor{gray}{464,330} & \textcolor{gray}{468,787} & \textcolor{gray}{590,908} \\
  & NSGA-II-100 & -- & -- & -- & -- & -- & -- & -- & -- & -- & -- \\
  & NSGA-II-500 & -- & -- & -- & -- & -- & -- & -- & -- & -- & -- \\
  & \noshRule{} & 50 & 50 & 50 & 50 & 50 & 5,000 & 5,000 & 5,000 & 5,000 & 5,000 \\
\midrule

\multirow{4}{*}{Time $\downarrow$} 
  & \textcolor{gray}{Exact} & \textcolor{gray}{1} & \textcolor{gray}{1} & \textcolor{gray}{2} & \textcolor{gray}{5} & \textcolor{gray}{31} & \textcolor{gray}{11} & \textcolor{gray}{51} & \textcolor{gray}{261} & \textcolor{gray}{567} & \textcolor{gray}{783} \\
  & NSGA-II-100 & 1 & 1 & 1 & 2 & 13 & 1 & 7 & 77 & 182 & 313 \\
  & NSGA-II-500 & 1 & 1 & 1 & 2 & 13 & 1 & 7 & 77 & 182 & 313 \\
  & \noshRule{} & 1 & 1 & 1 & 2 & 13 & 1 & 7 & 77 & 183 & 312 \\
\midrule

\multirow{4}{*}{Cardinality $\uparrow$} 
  & \textcolor{gray}{Exact} & \textcolor{gray}{100} & \textcolor{gray}{100} & \textcolor{gray}{100} & \textcolor{gray}{100} & \textcolor{gray}{100} & \textcolor{gray}{100} & \textcolor{gray}{100} & \textcolor{gray}{100} & \textcolor{gray}{100} & \textcolor{gray}{100} \\
  & NSGA-II-100 
    & 2 & 1 & 1 & 1 & 1
    & 0 & 0 & 0 & 0 & 0 \\
  & NSGA-II-500 
    & 0 & 0 & 0 & 0 & 4
    & 0 & 0 & 1 & 1 & 0 \\
  & \noshRule{} & \textbf{85} & \textbf{84} & \textbf{89} & \textbf{87} & \textbf{87} & \textbf{99} & \textbf{99} & \textbf{99} & \textbf{99} & \textbf{99} \\
\midrule

\multirow{4}{*}{Precision $\uparrow$} 
  & \textcolor{gray}{Exact} & \textcolor{gray}{100} & \textcolor{gray}{100} & \textcolor{gray}{100} & \textcolor{gray}{100} & \textcolor{gray}{100} & \textcolor{gray}{100} & \textcolor{gray}{100} & \textcolor{gray}{100} & \textcolor{gray}{100} & \textcolor{gray}{100} \\
  & NSGA-II-100 
    & 12 & 15 & 14 & 30 & 50
    & 0 & 8 & 24 & 31 & 38 \\
  & NSGA-II-500 
    & 0 & 0 & 0 & 1 & 56
    & 0 & 1 & 28 & 36 & 44 \\
  & \noshRule{} & \textbf{89} & \textbf{90} & \textbf{94} & \textbf{94} & \textbf{93} & \textbf{100} & \textbf{100} & \textbf{99} & \textbf{100} & \textbf{100} \\
\midrule

\multirow{4}{*}{IGD $\downarrow$} 
  & \textcolor{gray}{Exact} & \textcolor{gray}{0.000} & \textcolor{gray}{0.000} & \textcolor{gray}{0.000} & \textcolor{gray}{0.000} & \textcolor{gray}{0.000} & \textcolor{gray}{0.000} & \textcolor{gray}{0.000} & \textcolor{gray}{0.000} & \textcolor{gray}{0.000} & \textcolor{gray}{0.000} \\
  & NSGA-II-100 
    & 0.234 & 0.252 & 0.281 & 0.260 & 0.289
    & 0.437 & 0.205 & 0.232 & 0.293 & 0.309 \\
  & NSGA-II-500 
    & 1.413 & 1.234 & 1.111 & 0.487 & 0.198
    & 2.350 & 0.282 & 0.142 & 0.189 & 0.222 \\
  & \noshRule{} 
    & \textbf{0.012} & \textbf{0.016} & \textbf{0.013} & \textbf{0.017} & \textbf{0.019} 
    & \textbf{0.000} & \textbf{0.000} & \textbf{0.001} & \textbf{0.001} & \textbf{0.001} \\
\midrule

\multirow{4}{*}{$|\hat{\mathcal{Z}}^\star|$} 
  & \textcolor{gray}{Exact} & \textcolor{gray}{238} & \textcolor{gray}{1,117} & \textcolor{gray}{4,765} & \textcolor{gray}{9,117} & \textcolor{gray}{25,457} & \textcolor{gray}{787} & \textcolor{gray}{6,099} & \textcolor{gray}{29,061} & \textcolor{gray}{59,951} & \textcolor{gray}{103,489} \\
  & NSGA-II-100 
    & 26.7 & 45.8 & 63.8 & 96.6 & 100.0
    & 19.6 & 97.7 & 100.0 & 100.0 & 100.0 \\
  & NSGA-II-500 
    & 6.016 & 9.684 & 15.862 & 45.206 & 445.142
    & 0.078 & 74.084 & 499.028 & 499.981 & 499.956 \\
  & \noshRule{} & 226 & 1,051 & 4,591 & 8,550 & 24,534 & 786 & 6,089 & 28,953 & 59,724 & 103,275 \\
\midrule

Inst. &  & 100 & 100 & 100 & 100 & 100 & 100 & 100 & 92 & 54 & 27 \\  
\bottomrule
\end{tabular}}
    \label{tab:mis_result_complete}
\end{table}

\begin{table}[htbp!]
    \caption{MOKP results averaged across test instances. 
    NSGA-II-$p$ denotes NSGA-II with population size $p$.
    Refer to \Cref{sec:setup} for column description.}
    \centering
    \footnotesize
    \resizebox{0.7\linewidth}{!}{
    \begin{tabular}{rrlrrrrrrr}
    \toprule
    $N$ & $K$ & ~~Method & ~~Width & ~~Time $\downarrow$ & ~~Cardinality $\uparrow$ & ~~Precision $\uparrow$ & ~~IGD $\downarrow$ & ~~$|\hat{\mathcal{Z}}^\star|$ \\
    \midrule
    \multirow{8}{*}{40} & \multirow{8}{*}{7} 
    & \textcolor{gray}{Exact}        & \textcolor{gray}{9,709} & \textcolor{gray}{84} & \textcolor{gray}{100} & \textcolor{gray}{100} & \textcolor{gray}{0.000} & \textcolor{gray}{25,098} \\
    \cmidrule{3-9}
        & & NSGA-II-100 & -- & 60 & 0 & 0 & 0.282 & 100 \\
        & & NSGA-II-500 & -- & 60 & 0 & 0 & 0.199 & 500 \\
    \cmidrule{3-9}
        & & \multirow{2}{*}{\noshRule{}} 
            & 2,000 & \textbf{2} & 19 & 64 & 0.128 & 4,131 \\
        & &  
            & 3,000 & 19 & 61 & 89 & 0.047 & 12,972 \\
    \cmidrule{3-9}
        & & \multirow{2}{*}{\noshFE{}} 
            & 2,000 & 16 & 60 & 74 & 0.042 & 20,482 \\
        & &  
            & 3,000 & 36 & \textbf{88} & \textbf{96} & \textbf{0.012} & 22,267 \\
    
    \midrule
    
    \multirow{8}{*}{50} & \multirow{8}{*}{4} 
        & \textcolor{gray}{Exact}        & \textcolor{gray}{12,359} & \textcolor{gray}{7} & \textcolor{gray}{100} & \textcolor{gray}{100} & \textcolor{gray}{0.000} & \textcolor{gray}{3,564} \\
    \cmidrule{3-9}
        & & NSGA-II-100 & -- & 12 & 0 & 0 & 0.137 & 100 \\
        & & NSGA-II-500 & -- & 12 & 0 & 0 & 0.059 & 497 \\
    \cmidrule{3-9}
        & & \multirow{2}{*}{\noshRule{}} 
            & 2,500 & \textbf{1} & 17 & 36 & 0.085 & 1,132 \\
        & &  
            & 3,500 & 2 & 52 & 70 & 0.032 & 2,256 \\
    \cmidrule{3-9}
        & & \multirow{2}{*}{\noshFE{}} 
            & 2,500 & 3 & 61 & 70 & 0.020 & 3,062 \\
        & &  
            & 3,500 & 4 & \textbf{88} & \textbf{92} & \textbf{0.006} & 3,367 \\
    
    \midrule
    
    \multirow{8}{*}{80} & \multirow{8}{*}{3} 
        & \textcolor{gray}{Exact}        & \textcolor{gray}{20,097} & \textcolor{gray}{27} & \textcolor{gray}{100} & \textcolor{gray}{100} & \textcolor{gray}{0.000} & \textcolor{gray}{2,442} \\
    \cmidrule{3-9}
        & & NSGA-II-100 & -- & 58 & 0 & 0 & 0.068 & 100 \\
        & & NSGA-II-500 & -- & 58 & 0 & 0 & 0.023 & 499 \\
    \cmidrule{3-9}
        & & \multirow{2}{*}{\noshRule{}} 
            & 4,000 & \textbf{3} & 10 & 19 & 0.064 & 1,015 \\
        & &  
            & 6,000 & 7 & 62 & 71 & 0.013 & 1,954 \\
    \cmidrule{3-9}
        & & \multirow{2}{*}{\noshFE{}} 
            & 4,000 & 6 & 46 & 53 & 0.012 & 2,039 \\
        & &  
            & 6,000 & 12 & \textbf{93} & \textbf{95} & \textbf{0.002} & 2,363 \\
    \bottomrule
    \end{tabular}}
    \label{tab:kp_result_complete}
\end{table}

\end{appendix}

\newpage
\bibliography{references}

\end{document}